%% file: neurips_2025.tex
\documentclass{article}

\usepackage[preprint]{neurips_2025}

\usepackage[utf8]{inputenc} 
\usepackage[T1]{fontenc}    
\usepackage{hyperref}       
\usepackage{url}            
\usepackage{booktabs}       
\usepackage{amsfonts}       
\usepackage{nicefrac}       
\usepackage{microtype}      
\usepackage{xcolor}         

\usepackage{booktabs}
\usepackage{subcaption}
\usepackage{amsmath} 
\input{preamble}

\usepackage{graphicx}

\usepackage{algorithm}
\usepackage{algorithmic}
\usepackage{times}
\usepackage{mdframed}

\newmdenv[
  backgroundcolor=gray!15,
  linecolor=gray,
  leftmargin=0pt,
  rightmargin=0pt,
  innerleftmargin=6pt,
  innerrightmargin=6pt,
  innertopmargin=6pt,
  innerbottommargin=6pt,
  skipabove=6pt,
  skipbelow=6pt
]{neuripsbox}
\title{Tackling the Noisy Elephant in the Room: Label Noise-robust Out-of-Distribution Detection via Loss Correction and Low-rank Decomposition}
\workshoptitle{Tackling the Noisy Elephant in the Room: Label Noise-robust Out-of-Distribution Detection via Loss Correction and Low-rank Decomposition}

\author{%
  Tarhib Al Azad and Shahana Ibrahim\\
  Department of Electrical and Computer Engineering\\
  University of Central Florida\\
  Orlando, FL 32826 \\
}

\begin{document}

\maketitle

\begin{abstract}
  Robust out-of-distribution (OOD) detection is an indispensable component of modern artificial intelligence (AI) systems, especially in safety-critical applications where models must identify inputs from unfamiliar classes not seen during training. While OOD detection has been extensively studied in the machine learning literature—with both post hoc and training-based approaches—its effectiveness under noisy training labels remains underexplored. Recent studies suggest that label noise can significantly degrade OOD performance, yet principled solutions to this issue are lacking. 
In this work, we demonstrate that directly combining existing label noise-robust methods with OOD detection strategies is insufficient to address this critical challenge. To overcome this, we propose a \textit{robust} OOD detection framework that integrates loss correction techniques from the noisy label learning literature with low-rank and sparse decomposition methods from signal processing.  
Extensive experiments on both synthetic and real-world datasets demonstrate that our method significantly outperforms the state-of-the-art OOD detection techniques, particularly under severe noisy label settings.
\end{abstract}

\section{Introduction}
Artificial intelligence (AI) models have achieved remarkable performance across myrid of domains including computer vision and natural language processing. Yet, a persistent challenge arises in real-world deployment: these models often fail to recognize inputs from unfamiliar data distributions, leading to overly confident and potentially misleading predictions \cite{goodfellow2014explaining}. This limitation underscores the importance of out-of-distribution (OOD) detection for building trustworthy AI systems, particularly in high-stakes domains such as autonomous driving \cite{geiger2012we} and medical diagnostics \cite{thomas2017unsupervised}. The goal of OOD detection is not only to provide accurate prediction on seen data distributions but also to flag inputs from novel or unobserved distributions \cite{hendrycks2016baseline}.

{OOD} detection has been an active topic of research in the field of AI for many decades; a recent survey can be found in \cite{yang2024generalized}. A key focus in this field is detecting semantic shifts—scenarios where new, previously unseen classes appear in the test data, resulting in a mismatch between the label spaces of in-distribution (ID) and OOD samples. A wide range of methods have been proposed for OOD detection, including softmax/logit-based post-hoc techniques \cite{hendrycks2016baseline, liang2018enhancing,hendrycks2022scaling,sun2022dice,sun2021react,dong2022neural} and feature distance-based strategies \cite{lee2018simple,sun2022deepnn,ming2023hypersphericalood,SehwagSSD,ghosal2023overcomeood}. Nonetheless, most existing OOD detection methods are developed under the assumption that models are trained on clean, correctly labeled data. However, in practice, training datasets often contain noisy labels, stemming from the scarcity of expert annotators and the high cost of accurate label acquisition \cite{buhrmester2016amazon}. Recent empirical studies have brought serious attention to this issue, revealing that the presence of label noise can significantly degrade the performance of state-of-the-art OOD detection methods \cite{humblot2024noisy}. This highlights a critical gap in current research and underscores the need to develop robust OOD detection frameworks that remain reliable under real-world label noise.

The effect of label noise on the classification performance of the deep learning models has been extensively studied in recent years; see the survey \cite{song2022learning}. It is now well-established that training deep neural network (DNN) models with noisy labels can severely degrade classification performance, leading to poor generalization and overfitting \cite{arpit2017closer,zhang2017understanding}. To address this, a variety of label noise-robust methods have been proposed, including loss correction strategies such as probabilistic modeling techniques \cite{liu2016classification,patrini2017making,li2021provably,xia2020part, Yang2021EstimatingIB,cheng2020learning}, robust loss function designs \cite{zhang2018generalized,lyu2019curriculum,wang2019symmetric}, and in-built sample selection strategies \cite{jiang2018mentornet,yu2019does,nguyen2019self,han2018co, li2020dividemix}, 
However, their effectiveness in OOD detection under label noise remains largely unexplored. 
The key challenge lies in the misalignment of objectives: while label noise methods aim to correct the prediction probabilities within the training distribution, OOD detection requires learning discriminative feature representations to detect the samples that does not belong to the training distribution. Hence, most existing label-noise approaches exhibit poor OOD detection performance when applied directly, as we will demonstrate in detail in subsequent sections. 



\begin{figure}
    \centering
    \includegraphics[width=0.75\linewidth]{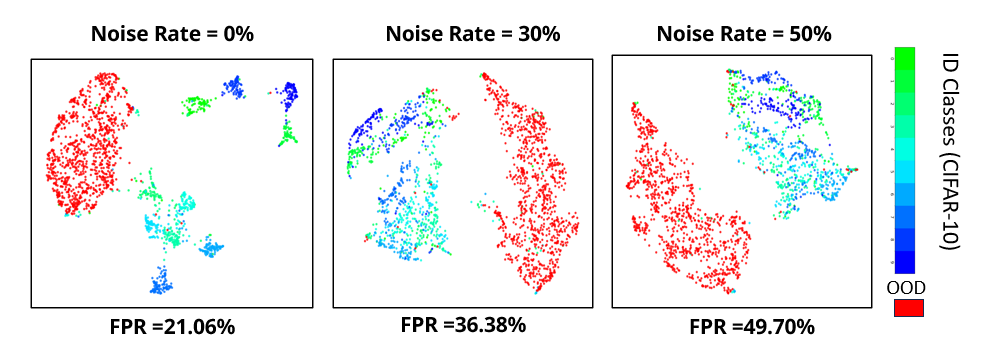}
    \caption{The effect of label noise for OOD detection. The figure shows the UMAP representations of the latent feature vectors $\bm h(\bm x)$ learned using the cross entropy loss-based training using the noisily labeled dataset $\{\bm x_n,\widehat{y}_n\}$ for various synthetic noise rates. The false positive ratio (FPR) for OOD detection using $k$NN score is also reported. The clusters are more distorted for the training data, losing the ID-ness characteristics, resulting in degraded performance in OOD detection during test time.}
    \label{fig:noise_knn}
\end{figure}

\noindent
\textbf{Our Contributions.} 
In this work, we investigate the critical challenge of robust OOD detection in the presence of noisy labels in the training set. Unlike existing studies that focus solely on the empirical limitations of current OOD detection methods \cite{humblot2024noisy}, we identify a key gap, where the label noise-robust methods improves generalization under noisy supervision for classification settings, yet are largely ineffective when directly applied for OOD detection. To address this limitation, 
we propose a novel learning framework, named as \underline{\textbf{N}}oise-robust \underline{\textbf{O}}ut-\underline{\textbf{O}}f-\underline{\textbf{D}}istribution \underline{\textbf{Le}}arning (NOODLE), by leveraging the loss correction techniques with low-rank and sparse decomposition methods.  To the best of our knowledge, this work is the first to offer a principled solution to the problem, achieving substantial improvements over state-of-the-art OOD detection methods in the presence of label noise. 



\noindent
{\bf Notation.} Notations are defined in the supplementary materials.

\section{Problem Statement}

 Consider an input feature space ${\mathcal{X}} \subset \mathbb{R}^D$, where $D$ denotes the dimensionality of the input features. Let the label space be defined as ${\mathcal{Y}} = \{1, \dots, K\}$, corresponding to $K$ distinct classes for the ID data.
We define the training dataset ${\mathcal{D}}$ as:
\begin{align*}
    {\mathcal{D}} = \{(\bm{x}_n, y_n)\}_{n=1}^N, \quad \bm{x}_n \in {\mathcal{X}}, \quad y_n \in {\mathcal{Y}},
\end{align*}
where $\bm{x}_n$ is the feature vector of the $n$-th training example, $y_n$ is its associated \textit{ground-truth} class label, and $N$ denotes the total number of training samples.
Each pair $(\bm{x}_n, y_n)$ is assumed to be drawn independently and identically distributed (i.i.d.) from an underlying joint distribution ${\mathcal{P}}_{{\mathcal{X}}{\mathcal{Y}}}$.
Let $\bm{h}: \mathbb{R}^D \rightarrow \mathbb{R}^L$ denote a DNN that maps each input $\bm{x}_n$ to an $L$-dimensional \textit{latent feature} representation $\bm{h}(\bm{x}_n)$. For the task of multi-class classification, we employ a projection head $\bm{c}: \mathbb{R}^L \rightarrow \mathbb{R}^K$ to produce pre-softmax logits. Thus, the overall label prediction function is given by:
\[
    \bm{f}(\bm{x}_n) = \bm \sigma(\bm{c}(\bm{h}(\bm{x}_n))),
\]
where $\bm \sigma$ denotes the softmax function that output the class probabilities. We often learn the parameters of these functions by training via cross-entropy (CE) minimization as follows:
\begin{align} \label{eq:ce}
    \text{minimize}_{\bm \theta}~ {\cal L}_{\text{CE}} (\bm \theta ; \{\bm x_n, y_n\}) = -\sum_{n=1}^N \sum_{k=1}^K \mathbb{I}[y_n=k]\log (\bm f(\bm{x}_n)))),
\end{align}
where $\bm \theta$ denotes the DNN parameters of both $\bm h$ and $\bm c$ functions.
 

\noindent
{\bf OOD Detection.} AI systems are generally learned under the closed-world assumption, where it is presumed that test samples are drawn from the same distribution as the training data. However, this assumption often fails in practical scenarios, where models inevitably encounter samples that lie outside the training distribution. These unfamiliar samples are known as OOD inputs~\cite{hendrycks2016baseline}.
In classification tasks, such distributional shift may manifest as a semantic shift, wherein some test instances originate from an \textit{unknown} label space ${\mathcal{Y}}^o$, disjoint from the known label space, i.e., ${\mathcal{Y}} \cap {\mathcal{Y}}^o = \emptyset$. The objective of OOD detection is to identify whether a given test input belongs to the in-distribution (ID) or not, thereby preventing the model from making confident predictions on OOD inputs.
Thus, OOD detection can be considered as a binary classification task that distinguishes ID samples from OOD ones. This can be formalized by a detection function:
\begin{align} \label{eq:detect}
    \bm{g}_{\tau}(\bm{x}) = 
    \begin{cases}
        {\sf ID} & \text{if } s(\bm{x}) \ge \tau, \\
        {\sf OOD} & \text{if } s(\bm{x}) < \tau,
    \end{cases}
\end{align}
where $s(\bm{x})$ is a scoring function that quantifies the likelihood of $\bm{x}$ belonging to the ID distribution, and $\tau$ is a predefined threshold. 

Typically, scoring function $s(\bm x)$ is derived from the trained parameters of the underlying DNN model. Several scoring functions have been proposed in the literature. Early OOD detection methods directly used the softmax outputs $\bm f(\bm x)$ to score "OOD-ness" \cite{hendrycks2016baseline, liang2018enhancing}, but they suffered from overconfidence issues, reducing the desired ID-OOD separability. Further, pre-softmax activations-based approaches (e.g., by using the logits $\bm c(\bm h(\bm x))$) were introduced \cite{hendrycks2022scaling,sun2022dice,sun2021react,dong2022neural}, though they remained sensitive to architecture and still faced overconfidence issues. Recently, distance-based methods such as those based on Mahalanobis \cite{lee2018simple} and $k$-nearest neighbor ($k$NN) \cite{sun2022deepnn,ming2023hypersphericalood,SehwagSSD,ghosal2023overcomeood} have gained traction by leveraging the clusterability  of latent feature representations $\bm h(\bm x)$.
In essence, the success of OOD detection lies in the careful design and learning of the scoring function $\bm s(\bm x)$ that can ensure the ID-OOD separability during test time.

\noindent
{\bf Learning under Label Noise.} Most studies in the domain of OOD detection assume that the DNN classifier $\bm f$ and the scoring function $\bm s$ are learned using ground-truth labels \( y_n \). However, the lack of access to reliable ground-truth annotations is a significant challenge for robust OOD detection--see an example in Fig. \ref{fig:noise_knn} where the clusterability of the latent representations $\bm{h}(\bm{x})$ is severely compromised under label noise, leading to significant degradation in ID-ODD separability for the $k$NN score function. 

In scenarios where ground-truth labels $y_n$ are difficult to obtain, we often rely on their noisy counterparts, denoted by \( \widehat{y}_n \in \{1, \dots, K\} \), associated with each data item \( \bm x_n \). In noisy label settings, for many data items, the observed label does not match the true label, i.e., \( \widehat{y}_n \ne y_n \). 

\begin{mdframed}[backgroundcolor=gray!15, linecolor=gray]
The goal of \textit{label noise-robust OOD detection} is two-fold: \textit{(i)} accurately classify ID samples through a well-generalized predictor $\bm{f}$, and \textit{(ii)} reliably detect OOD instances using a robust decision function $\bm{g}_\tau$, despite learning them using the noisily labeled dataset 
$
    {\widehat{\mathcal{D}}} = \{(\bm{x}_n, \widehat{y}_n)\}_{n=1}^N, ~\bm{x}_n \in {\mathcal{X}}, ~ \widehat{y}_n \in {\mathcal{Y}}$.
    
\end{mdframed}





\section{Proposed Approach}

In this section, we present our label noise-robust OOD detection framework. Our strategy is based on \textit{cleansing} the noise-corrupted latent feature space using an end-to-end training strategy, thereby making it robust for OOD detection at test time. Towards this goal, our framework encompasses three main components: \textit{i)} loss correction module \textit{ii)} low rank and sparse decomposition of latent feature matrix \textit{iii)} OOD detection using distance-based metrics, e.g., $k$NN.

\subsection{Loss Correction}
As demonstrated in Fig. \ref{fig:noise_knn}, training directly with noisy labels (e.g., by using the CE minimization as in \eqref{eq:ce} where the unobserved ground-truth labels $y_n$ are replaced by the observed noisy labels $\widehat{y}_n$) leads to a corrupted latent feature space. To address this, we first integrate a loss correction module to reduce the effect of label noise in learned features. Loss correction strategies have attracted considerable attention in noisy label learning literature. Among these, probabilistic noise modeling via the so-called \textit{transition matrices} \cite{patrini2017making,tanno2019learning,li2021provably,ibrahim2023deep} and robust loss function-based approaches \cite{zhang2018generalized,lyu2019curriculum,wang2019symmetric} are particularly well-received,
owing to their strong theoretical foundations and robust empirical performance in classification tasks. 

In general, loss correction strategies design a modified cross-entropy loss to train the classifier $\bm f$ on the noisy dataset $\{\bm x_n,\widehat{y}_n\}$, while aiming to predict the ground-truth labels, i.e., 
\begin{align} \label{ee:rob_loss}
    \underset{\bm{\theta},\, \bm{\eta} }{\text{minimize}} ~{\cal L}_{\text{CE}}^{\text{mod}}(\bm \theta, \bm \eta; \{\bm x_n, \widehat{y}_n\})
\end{align}
where $\bm{\eta}$ typically refers to additional model parameters according to specific loss designs. 
For instance, in the case of transition matrix-based approaches \cite{patrini2017making,tanno2019learning,li2021provably,ibrahim2023deep}, $\bm{\eta}$ refers to the noise transition probabilities that learns the probabilistic label confusion terms ${\sf Pr}(\widehat{y}_n=k |y_n=k')$. In sample selection approaches \cite{jiang2018mentornet,yu2019does,nguyen2019self,han2018co, li2020dividemix} , $\bm \eta$ instead represents sample-weighting terms that regulate the contributions of clean and noisy sample–label pairs.
In contrast, robust loss function-based methods, e.g., \cite{zhang2018generalized,wang2019symmetric}, often do not introduce additional parameters---they directly design loss functions that are inherently less sensitive to incorrect labels. For instance, symmetric cross-entropy (SCE) \cite{wang2019symmetric} and generalized cross-entropy (GCE) \cite{zhang2018generalized} can be viewed as hybrids of CE loss and mean absolute error (MAE) loss, thereby combining the favorable convergence properties of CE with the robustness of MAE against outliers.


Nonetheless, these loss correction strategies  primarily operate by modifying the softmax prediction outputs of the ID samples rather than directly \textit{correcting} their feature embeddings. However, feature embeddings are often more critical for OOD detection, particularly for the competitive, distance-based OOD metrics such as 
$k$-nearest neighbor \cite{sun2022deepnn} and Mahalanobis \cite{lee2018simple}. This misalignment of objectives results in suboptimal performance in mitigating the effect of label noise in OOD detection.

\subsection{Low-rank plus Sparse Decomposition}
To overcome the limitation of loss correction modules in handling {feature correction}, we introduce the next key component of our framework. A critical observation underlying its design is that, in the absence of label noise, latent feature vectors naturally exhibit certain clustering patterns, reflecting their low-rank structure due to their class-specific organization---see the first UMAP plot in Fig. \ref{fig:noise_knn}. This intrinsic structural tendency can be explicitly leveraged in the training phase to encourage low-rank properties in the feature representations. To this end, we adopt a low-rank and sparse decomposition strategy, drawing inspiration from classical signal processing techniques \cite{candes2011robust,zhang2011image}.

Consider the latent feature representation $\bm h(\bm x)$ of the input image $\bm x$ (e.g., the penultimate layer encoding of the DNN model). Let us represent the latent feature matrix $\bm H$ as follows:
\begin{equation}
    \bm{H} = \big[\,\bm{h}(\bm{x}_1), \dots, \bm{h}(\bm{x}_N)\,\big] \in \mathbb{R}^{D \times N}
    \label{eq:feature-matrix}
\end{equation}

where $D$ is the feature dimension and $N$ is the batch size. In order to exploit the low-rankness of the latent matrix  $\bm H$ along with a sparse structure, we assume that $\bm H \approx \bm L +\bm S,$ where $\bm L \in \mathbb{R}^{ D \times N}$ is the low-rank component and $\bm S \in \mathbb{R}^{D \times N}$ is a column sparse matrix, i.e., most columns of $\bm S$ has zero $\ell_2$ norm. That means, the low-rank term captures the underlying class structure information, whereas the sparse term can handle the outlier data items that does not strictly conform to the low-rank assumption.

Learning $\bm L$ and $\bm S$ from the observed matrix $\bm H$ generally involves solving optimization problem of the form \cite{candes2009rpca,alternateminialgo}:
%
\[
\min_{\bm L, \bm S} \ \|\bm L\|_* + \lambda \|\bm S\|_{2,1} \quad \text{s.t.} \quad \bm H = \bm L + \bm S,
\]
where $\|\bm L\|_*$ denotes the nuclear norm of $\bm L$ to promote the low-rankness and $\|\bm S\|_{2,1}$ denotes the matrix mixed norm that promotes column sparsity in $\bm S$. Here, $\lambda > 0$ is a regularization parameter that balances the contributions of the low-rank and sparse terms. As computing the nuclear norm involves costly operations like singular value decomposition, we adopt an efficient power iteration (PI)-based low-rank approximation strategy \cite{rokhlin2010randomized,gu2015subspace} in our training phase. Specifically, The method estimates the top-$K$ left singular vectors of the latent representation $\bm H$ by iteratively refining a randomly initialized orthonormal basis $\bm Q \in \mathbb{R}^{D \times K}$ through alternating projections of $\bm H$ and orthonormalization via QR decomposition. Here, The rank $K$ can be selected according to the number of classes (or based on the number of coarse-grained classes in the case of datasets with very large label space).  Using the learned $\bm Q$, we decompose the feature matrix as
\[ 
\bm H_{\text{ID}} = \begin{bmatrix} \bm h_{\text{ID}}(\bm x_1), \dots, \bm h_{\text{ID}}(\bm x_N) \end{bmatrix} = (\bm Q \bm Q^\top) \bm H, ~~
\bm H_{\text{OOD}} = \begin{bmatrix} \bm h_{\text{OOD}}(\bm x_1), \dots, \bm h_{\text{OOD}}(\bm x_N) \end{bmatrix}= \bm H - \bm H_{\text{ID}},
\]
where $\bm H_{\text{ID}} \in \mathbb{R}^{D \times N}$  represents the ID component and $\bm H_{\text{OOD}} \in \mathbb{R}^{D \times N}$ contains residual features that may potentially carry non-ID information. Further, to enforce the column sparsity in the matrix $\bm H_{\text{OOD}}$, we employ the following regularization term:
\begin{equation}
\label{eq:col_sparsity}
\mathcal{L}_{\text{sparse}} = \|\bm H_{\text{OOD}}\|_{2,1}
= \sum_{j=1}^N \sqrt{\sum_{i=1}^D \left(\bm H_{\text{OOD}} \right)_{ij}^2}.
\end{equation}


Finally, the proposed method is trained by minimizing a joint objective that combines the modified cross-entropy loss as explained in \eqref{ee:rob_loss} with the regularizer in \eqref{eq:col_sparsity}:
\[
\mathcal{L}_{\mathcal{F}}
= \mathcal{L}_{\text{CE}}^{\text{mod}}
+ \lambda \, \mathcal{L}_{\text{sparse}},
\]
where $\lambda > 0$ is a regularization hyperparameter that controls the strength of the column-sparsity term. The detailed algorithm is presented in the supplementary section.

\subsection{OOD Detection with Refined Feature Representations}
After training, we adopt a distance-based approach for OOD detection using the cleaned latent features $\bm h_{\text{ID}}(\bm x)$. 
Towards this, we extract the $\ell_2$-normalized feature vectors $\bm{u}_{\text{ID}}(\bm x_n) = \bm h_{\text{ID}}(\bm x_n) / \|\bm h_{\text{ID}}(\bm x_n)\|_2$ for all ID training samples and store them as reference embeddings. 
At test time, a query sample $\bm{x}^*$ is mapped to its normalized feature $\bm{u}(\bm x^*) = \bm{h}(\bm{x}^*) / \|\bm{h}(\bm{x}^*)\|_2$, whose distance to the stored ID embeddings $\bm{u}_{\text{ID}}(\bm x_n)$ is then evaluated. Following prior work, we adopt different distance metrics, such as $k$-nearest neighbor \cite{sun2022deepnn} and Mahalanobis distance \cite{lee2018simple}. To be specific, in the case of $k$-nearest neighbor metric, 
we select the $k$-th smallest distance to define the OOD score $s(\bm{x})$ (also see \eqref{eq:detect}):
\[
s_{\text{kNN}}(\bm{x}^*) = -\big\| \bm{u}(\bm x^*) - \bm{u}^{(k)}_{\text{ID}} \big\|_2,
\]
where $\bm{u}_{\text{ID}}^{(k)}$ denotes the $k$-th nearest neighbor embedding from the cleaned latent features of the training data. 
A decision threshold $\tau$ of the detection function $g_{\tau}$ is chosen based on a validation set such that a high fraction (e.g., 95\%) of ID samples are correctly classified as ID.

\section{Experiments}
In this section, we present a series of experiment results to showcase the effectiveness of our label noise-robust OOD detection framework.

\textbf{Datasets.} For synthetic label noise settings, we consider CIFAR-10~\cite{Krizhevsky2009LearningML} as ID dataset. 
CIFAR-10 consists of 50{,}000 training images and 10{,}000 test images across 10 different classes.  For synthetic label noise generation, we adopt class-independent symmetric noise, where every true label has the same probability 
 of being corrupted, and when corrupted, it is flipped uniformly at random to any of the other $K-1$ classes, regardless of the original class. We vary the noise rate at 10\%, 30\%, and 50\% to simulate different levels of noise severity.

To test under realistic label noise, we also consider the human-annotated noisy label datasets CIFAR-10N, CIFAR-100N~\cite{wei2022learning}, and Animal-10N~\cite{song2019selfie}. These are annotated by the crowd workers from the popular crowdsourcing platform Amazon Mechanical Turk (AMT). CIFAR-10N provides five types of noisy label sets: \textit{worst}, \textit{aggregate}, \textit{random1}, \textit{random2}, and \textit{random3}, while we use the \textit{fine} type label noise for CIFAR-100N. Similar to CIFAR-10, Animal-10N contains 50{,}000 training and 5{,}000 test images across 10 classes, with human-annotated noisy labels. CIFAR-100N contains the same number of images but is divided into 100 fine-grained classes, making both classification and OOD detection more challenging. As OOD datasets during test time, we consider several benchmark datasets, including SVHN~\cite{netzer}, FashionMNIST~\cite{XiaoFahionmnist}, LSUN~\cite{yu2016lsun}, iSUN~\cite{panisun}, Texture~\cite{Cimpoi}, and Places365~\cite{zhou2016}.

\noindent
{\bf Baselines.} We compare our proposed method with several OOD detection baselines as well as different label noise-robust techniques.

Regarding the OOD detection baselines, we consider MSP \cite{hendrycks2017baseline}, ODIN \cite{liang2018enhancing}, Energy \cite{Liu2020EnergybasedOD}, ReAct \cite{sun2021react}, Mahalanobis \cite{lee2018simple}, KNN \cite{sun2022deepnn}, CIDER \cite{ming2023hypersphericalood}, SSD+ \cite{SehwagSSD}, and SNN \cite{ghosal2023overcomeood}. MSP, ODIN and Energy are softmax-based approaches.
 MSP relies only on softmax output of the model, while ODIN uses an additional temperature scaling hyperparameter. Energy method computes an energy-based metric from the model outputs, identifying test samples with higher energy as OOD. ReAct is a logit-based approach. Mahalanobis, KNN, CIDER, SSD+, and SNN are distance-based approaches. For MSP, ODIN, Energy, ReAct, and SNN, the DNN encoder is trained using the standard cross-entropy loss. For KNN and SSD+, supervised contrastive loss \cite{khosla2020supervised} is used. CIDER is trained using a maximum likelihood estimation-based loss together with dispersion regularization. As previously discussed, most recent distance-based methods—such as KNN, CIDAR, and SNN all advocate the use of non-parametric $k$NN-based score \cite{sun2022deepnn} for OOD detection.


{Regarding label-noise-robust learning methods, we consider different lines of approach for our evaluation. Specifically, we consider CM \cite{Li2021ProvablyEL}, VolMinNet \cite{Li2021ProvablyEL}, SCE \cite{wang2019symmetriccrossentropyrobust}, GCE \cite{zhang2018generalizedcrossentropyloss}, DivideMix \cite{li2020dividemixlearningnoisylabels}, and Co-Teaching \cite{han2018coteachingrobusttrainingdeep}. Here, CM and VolMinNet are probabilistic noise-modeling approaches that rely on transition matrices to correct label noise. GCE and SCE are robust loss function–based approaches that are variants of the cross-entropy loss. DivideMix and Co-Teaching are sample-selection-based approaches that focus on reweighting samples based on the presence of label noise. Here, DivideMix identifies small-loss (likely clean) samples and applies semi-supervised learning to the noisy labeled samples, while Co-Teaching uses two networks that are trained simultaneously and exchange small-loss samples with each other. For OOD performance evaluation, we use the $k$NN-based metric for all these methods, unless specified otherwise. }

\begin{table*}[t]
\centering
\caption{OOD detection performance on CIFAR10 with synthetic label noise across different OOD datasets; The top two performing algorithms (in terms of average FPR95) are highlighted in bold.}
\label{tab:cifar10n_symmetric}
\setlength{\tabcolsep}{3pt} 
\scriptsize
\resizebox{\textwidth}{!}{ 
\begin{tabular}{lcccccccccccccc}
\toprule
\textbf{Method} & \multicolumn{2}{c}{\textbf{SVHN}} & \multicolumn{2}{c}{\textbf{FashionMNIST}} & \multicolumn{2}{c}{\textbf{LSUN}} & \multicolumn{2}{c}{\textbf{iSUN}} & \multicolumn{2}{c}{\textbf{DTD/Texture}} & \multicolumn{2}{c}{\textbf{Places365}} & \multicolumn{2}{c}{\textbf{Average}} \\
\cmidrule(lr){2-3}\cmidrule(lr){4-5}\cmidrule(lr){6-7}\cmidrule(lr){8-9}\cmidrule(lr){10-11}\cmidrule(lr){12-13}\cmidrule(lr){14-15}
& FPR95$\downarrow$ & AUROC$\uparrow$ 
& FPR95 & AUROC 
& FPR95 & AUROC 
& FPR95 & AUROC 
& FPR95 & AUROC 
& FPR95 & AUROC 
& FPR95 & AUROC \\
\midrule
\multicolumn{15}{c}{\textit{Noise rate = 10\%}} \\
\texttt{KNN} & $34.23$ & $93.92$ & $38.51$ & $93.57$ & $17.71$ & $96.66$ & $21.72$ & $95.93$ & $28.58$ & $94.04$ & $62.26$ & $84.62$ & $33.84$ & $93.12$ \\
\texttt{MSP} & $73.54$ & $84.82$ & $54.92$ & $88.64$ & $32.84$ & $94.49$ & $55.18$ & $89.70$ & $77.02$ & $75.86$ & $69.01$ & $80.33$ & $60.42$ & $85.64$ \\
\texttt{ODIN} & $87.82$ & $60.19$ & $64.22$ & $79.77$ & $24.42$ & $94.56$ & $32.99$ & $91.49$ & $80.30$ & $56.06$ & $78.84$ & $66.87$ & $61.43$ & $74.82$ \\
\texttt{Energy} & $80.31$ & $77.93$ & $57.89$ & $84.84$ & $17.74$ & $96.65$ & $54.63$ & $88.38$ & $82.43$ & $62.42$ & $78.29$ & $72.39$ & $60.37$ & $79.12$ \\
\texttt{ReAct} & $96.77$ & $53.22$ & $63.00$ & $87.87$ & $49.38$ & $90.45$ & $72.98$ & $81.08$ & $92.89$ & $44.08$ & $75.21$ & $75.29$ & $75.04$ & $72.00$ \\
\texttt{Mahalanobis} & $31.71$ & $91.10$ & $72.56$ & $74.47$ & $28.29$ & $93.87$ & $52.77$ & $81.81$ & $49.11$ & $80.53$ & $94.74$ & $44.55$ & $54.86$ & $77.72$ \\
\texttt{CIDER} & $99.64$ & $51.13$ & $99.90$ & $27.39$ & $99.81$ & $24.37$ & $99.84$ & $27.25$ & $93.72$ & $39.31$ & $100.00$ & $9.42$ & $98.80$ & $29.83$ \\
\texttt{SSD+} & $90.96$ & $73.67$ & $98.72$ & $46.18$ & $99.69$ & $40.58$ & $100.00$ & $26.23$ & $98.06$ & $33.48$ & $99.15$ & $37.90$ & $97.76$ & $43.01$ \\
\texttt{SNN} & $49.71$ & $91.73$ & $29.72$ & $95.06$ & $20.96$ & $96.25$ & $25.36$ & $95.23$ & $34.26$ & $92.53$ & $56.29$ & $86.46$ & $36.05$ & $92.88$ \\

\texttt{SCE} & $5.97$ & $98.96$ & $12.60$ & $97.80$ & $2.93$ & $99.44$ & $15.62$ & $97.01$ & $27.16$ & $94.28$ & $59.64$ & $85.79$ & $20.66$ & $95.55$ \\
\texttt{GCE} & $7.81$ & $98.50$ & $16.20$ & $97.33$ & $5.88$ & $98.96$ & $12.03$ & $97.92$ & $31.63$ & $93.56$ & $47.83$ & $89.35$ & $20.23$ & $95.94$ \\
\texttt{Co-teaching} & $40.12$ & $90.87$ & $99.29$ & $59.93$ & $75.83$ & $76.34$ & $96.95$ & $54.21$ & $49.45$ & $82.48$ & $93.19$ & $48.57$ & $75.81$ & $68.73$ \\
\texttt{DivideMix} & $62.65$ & $87.37$ & $68.10$ & $81.99$ & $49.01$ & $91.24$ & $42.84$ & $91.52$ & $37.48$ & $92.79$ & $77.16$ & $75.03$ & $56.21$ & $86.66$ \\
\texttt{CM} & $6.14$ & $98.90$ & $15.80$ & $97.05$ & $6.03$ & $98.91$ & $10.40$ & $98.12$ & $19.08$ & $96.47$ & $53.54$ & $87.17$ & $18.50$ & $96.10$ \\

\texttt{VolMinNet} & $2.64$ & $99.47$ & $5.18$ & $98.94$ & $5.00$ & $98.98$ & $9.26$ & $98.24$ & $21.60$ & $95.86$ & $56.16$ & $87.43$ & $\textbf{16.64}$ & $96.49$ \\
\midrule
\texttt{NOODLE} & $3.51$ & $ 99.28$ & $5.03$ & $98.97$ & $3.31  $ & $99.33$ & $3.05 $ & $99.22$ & $16.61$ & $96.78$ & $48.02$ & $89.62$ & $\textbf{13.26}$ & $97.20$ \\

\midrule

\multicolumn{15}{c}{\textit{Noise rate = 30\%}} \\
\texttt{KNN} & $23.80$ & $95.80$ & $36.15$ & $93.43$ & $27.04$ & $94.61$ & $22.03$ & $95.86$ & $39.50$ & $90.40$ & $69.76$ & $83.53$ & $36.38$ & $92.27$ \\
\texttt{MSP} & $76.88$ & $80.34$ & $56.16$ & $87.14$ & $29.90$ & $93.92$ & $58.82$ & $88.30$ & $79.04$ & $71.54$ & $74.98$ & $76.61$ & $62.63$ & $82.97$ \\
\texttt{ODIN} & $83.79$ & $61.60$ & $50.70$ & $83.57$ & $22.26$ & $94.83$ & $35.64$ & $89.98$ & $79.68$ & $54.97$ & $83.19$ & $60.62$ & $59.21$ & $74.26$ \\
\texttt{Energy} & $76.67$ & $76.82$ & $51.45$ & $86.81$ & $19.38$ & $95.65$ & $66.77$ & $84.97$ & $80.89$ & $62.95$ & $77.32$ & $72.32$ & $62.08$ & $79.92$ \\
\texttt{ReAct} & $88.89$ & $67.41$ & $62.61$ & $85.50$ & $21.63$ & $95.34$ & $90.19$ & $65.23$ & $91.33$ & $51.51$ & $82.51$ & $68.95$ & $72.86$ & $72.33$ \\
\texttt{Mahalanobis} & $37.86$ & $90.25$ & $50.42$ & $85.21$ & $26.98$ & $93.29$ & $60.92$ & $79.60$ & $52.75$ & $77.45$ & $95.94$ & $40.86$ & $54.15$ & $77.78$ \\
\texttt{CIDER} & $99.64$ & $51.13$ & $99.90$ & $27.39$ & $99.81$ & $24.37$ & $99.84$ & $27.25$ & $93.72$ & $39.31$ & $100.00$ & $9.42$ & $98.82$ & $29.81$ \\
\texttt{SSD+} & $91.27$ & $73.92$ & $98.72$ & $46.08$ & $99.69$ & $40.36$ & $100.00$ & $26.00$ & $98.06$ & $33.49$ & $99.15$ & $37.93$ & $97.82$ & $42.96$ \\
\texttt{SNN} & $23.37$ & $95.72$ & $34.55$ & $94.11$ & $25.91$ & $94.90$ & $34.38$ & $92.36$ & $42.27$ & $89.39$ & $65.06$ & $84.19$ & $37.59$ & $91.78$ \\

\texttt{SCE} & $19.48$ & $96.45$ & $25.84$ & $95.37$ & $16.58$ & $96.57$ & $61.71$ & $87.42$ & $35.50$ & $91.76$ & $74.66$ & $79.27$ & $38.96$ & $91.14$ \\
\texttt{GCE} & $58.38$ & $91.40$ & $20.37$ & $96.64$ & $11.32$ & $97.95$ & $12.38$ & $97.73$ & $30.53$ & $94.08$ & $51.59$ & $88.18$ & $30.76$ & $94.33$ \\
\texttt{Co-teaching} & $50.10$ & $83.16$ & $99.99$ & $20.37$ & $96.73$ & $64.14$ & $97.86$ & $41.70$ & $53.71$ & $79.45$ & $93.84$ & $48.63$ & $82.04$ & $56.24$ \\
\texttt{DivideMix} & $58.39$ & $90.07$ & $31.17$ & $94.93$ & $27.86$ & $95.59$ & $16.38$ & $96.93$ & $36.28$ & $92.76$ & $59.28$ & $84.34$ & $38.22$ & $92.44$ \\
\texttt{CM} & $22.04$ & $96.76$ & $8.79$ & $98.04$ & $10.17$ & $98.12$ & $23.30$ & $95.64$ & $23.71$ & $94.99$ & $55.42$ & $86.90$ & $\textbf{23.90}$ & $95.08$ \\

\texttt{VolMinNet} & $4.99$ & $99.04$ & $14.01$ & $97.09$ & $9.48$ & $98.33$ & $51.23$ & $89.68$ & $27.84$ & $93.44$ & $59.25$ & $85.42$ & $27.80$ & $93.84$ \\
\midrule
\texttt{NOODLE} & $1.84$ & $99.60$ & $19.66$ & $96.36$ & $7.28$ & $95.53$ & $10.76$ & $97.89$ & $20.67$ & $95.85$ & $57.50$ & $85.87$ & $\textbf{19.62}$ & $95.68$ \\
\midrule

\multicolumn{15}{c}{\textit{Noise rate = 50\%}} \\
\texttt{KNN} & $65.53$ & $85.64$ & $37.84$ & $93.71$ & $30.61$ & $93.38$ & $45.41$ & $89.21$ & $43.81$ & $89.06$ & $74.98$ & $79.58$ & $49.70$ & $88.43$ \\
\texttt{MSP} & $96.92$ & $53.68$ & $80.68$ & $77.70$ & $47.78$ & $89.67$ & $67.84$ & $83.23$ & $82.50$ & $68.45$ & $81.12$ & $73.63$ & $76.14$ & $74.40$ \\
\texttt{ODIN} & $94.94$ & $44.15$ & $71.46$ & $80.28$ & $34.04$ & $91.82$ & $47.51$ & $88.11$ & $79.61$ & $60.80$ & $82.87$ & $66.17$ & $68.40$ & $71.89$ \\
\texttt{Energy} & $97.93$ & $46.79$ & $83.41$ & $76.67$ & $39.77$ & $90.87$ & $67.60$ & $81.00$ & $85.43$ & $60.47$ & $82.28$ & $69.91$ & $76.07$ & $70.95$ \\
\texttt{ReAct} & $99.19$ & $24.75$ & $90.11$ & $64.20$ & $50.47$ & $85.45$ & $78.12$ & $67.79$ & $93.79$ & $39.56$ & $86.40$ & $62.25$ & $83.02$ & $57.33$ \\
\texttt{Mahalanobis} & $55.77$ & $83.12$ & $59.93$ & $85.98$ & $31.23$ & $93.54$ & $45.17$ & $88.63$ & $48.90$ & $81.70$ & $93.28$ & $51.47$ & $55.71$ & $80.74$ \\
\texttt{CIDER} & $99.65$ & $51.22$ & $99.91$ & $27.39$ & $100.00$ & $9.42$ & $99.84$ & $27.25$ & $93.72$ & $39.31$ & $99.81$ & $24.37$ & $98.82$ & $29.83$ \\
\texttt{SSD+} & $91.35$ & $74.03$ & $98.75$ & $46.17$ & $99.74$ & $40.57$ & $100.00$ & $25.89$ & $98.06$ & $33.47$ & $99.15$ & $37.91$ & $97.84$ & $43.01$ \\
\texttt{SNN} & $71.41$ & $83.96$ & $68.22$ & $87.66$ & $53.56$ & $89.27$ & $63.49$ & $80.99$ & $56.95$ & $85.48$ & $82.28$ & $77.12$ & $65.99$ & $84.08$ \\

\texttt{SCE} & $14.10$ & $97.40$ & $42.30$ & $90.96$ & $25.18$ & $94.09$ & $67.17$ & $80.81$ & $51.51$ & $84.65$ & $70.67$ & $77.75$ & $45.15$ & $87.61$ \\
\texttt{GCE} & $19.19$ & $96.43$ & $29.10$ & $95.06$ & $22.98$ & $95.40$ & $53.92$ & $86.31$ & $48.35$ & $87.54$ & $65.83$ & $83.58$ & $39.89$ & $90.72$ \\
\texttt{Co-teaching} & $57.05$ & $76.43$ & $99.97$ & $28.21$ & $99.23$ & $55.11$ & $96.99$ & $52.50$ & $54.45$ & $78.87$ & $94.22$ & $47.36$ & $83.65$ & $56.41$ \\
\texttt{DivideMix} & $24.69$ & $95.75$ & $40.94$ & $93.32$ & $37.02$ & $94.36$ & $20.81$ & $96.10$ & $53.10$ & $89.35$ & $56.39$ & $86.71$ & $38.82$ & $92.60$ \\
\texttt{CM} & $17.37$ & $96.91$ & $21.93$ & $95.56$ & $17.16$ & $96.58$ & $39.52$ & $92.86$ & $30.23$ & $93.36$ & $61.88$ & $84.82$ & $\textbf{31.35}$ & $93.35$ \\

\texttt{VolMinNet} & $13.01$ & $97.74$ & $15.36$ & $97.11$ & $14.18$ & $97.24$ & $60.13$ & $80.22$ & $45.85$ & $87.40$ & $55.26$ & $86.94$ & $33.96$ & $91.11$ \\
\midrule
\texttt{NOODLE} & $6.35$ & $98.43$ & $17.83$ & $96.58$ & $7.09$ & $98.50$ & $32.28$ & $93.92$ & $30.09$ & $92.47$ & $70.41$ & $81.24$ & $\textbf{27.34}$ & $93.52$ \\
\bottomrule
\end{tabular}
}
\end{table*}

\begin{table*}[t]
\centering
\caption{Average OOD detection performance on noisy real datasets; The top two performing algorithms (in terms of average FPR95) are highlighted in bold.}
\label{tab:cifar10n-animal10n}
\setlength{\tabcolsep}{3.5pt}
\renewcommand{\arraystretch}{1.2} 
\resizebox{\textwidth}{!}{ 
\begin{tabular}{lcccccccccccccccc}
\toprule
\textbf{Method}
& \multicolumn{12}{c}{\textbf{CIFAR-10N}} 
& \multicolumn{2}{c}{\textbf{Animal-10N}} &  \multicolumn{2}{c}{\textbf{CIFAR-100N}} \\
\cmidrule(lr){2-13}\cmidrule(lr){14-15}\cmidrule(lr){16-17}
& \multicolumn{2}{c}{Clean} 
& \multicolumn{2}{c}{Worst} 
& \multicolumn{2}{c}{Agg} 
& \multicolumn{2}{c}{Rand1} 
& \multicolumn{2}{c}{Rand2} 
& \multicolumn{2}{c}{Rand3}
& FPR95 & AUROC & FPR95 & AUROC \\
& FPR95 & AUROC 
& FPR95 & AUROC 
& FPR95 & AUROC 
& FPR95 & AUROC 
& FPR95 & AUROC 
& FPR95 & AUROC 
&       &       & &\\
\midrule
\texttt{KNN}              & $21.06$ & $95.80$ & $32.48$ & $92.89$ & $23.95$ & $94.84$ & $35.48$ & $92.65$ & $31.99$ & $92.70$ & $27.27$ & $94.09$ & $70.44$ & $77.04$ & ${43.20}$ & $86.54$\\
\texttt{MSP}              & $56.43$ & $90.07$ & $60.15$ & $85.49$ & $55.04$ & $88.21$ & $60.75$ & $86.62$ & $56.44$ & $86.50$ & $53.90$ & $86.70$ & $90.64$ & $59.90$ & $81.08$ & $72.66$\\
\texttt{ODIN}             & $33.10$ & $92.47$ & $45.31$ & $86.69$ & $43.83$ & $89.03$ & $49.12$ & $86.56$ & $46.71$ & $84.24$ & $41.91$ & $87.37$ & $76.97$ & $62.55$ & $71.72$ & $76.64$\\
\texttt{Energy}           & $39.15$ & $92.03$ & $47.30$ & $87.75$ & $56.03$ & $87.90$ & $54.89$ & $87.24$ & $50.31$ & $86.11$ & $42.74$ & $88.90$ & $75.60$ & $74.52$ & $78.93$ & $51.74$\\
\texttt{ReAct}            & $60.31$ & $83.12$ & $65.89$ & $78.86$ & $47.17$ & $91.01$ & $68.77$ & $76.92$ & $65.46$ & $79.83$ & $57.81$ & $81.24$ & $79.00  $ & $71.15$ & $76.24$ & $67.63$\\
\texttt{Mahalanobis}      & $47.22$ & $82.91$ & $53.57  $ & $80.61$ & $51.14  $ & $81.76$ & $55.26$ & $80.05$ & $44.25 $ & $ 84.60 $ & $48.37$ & $83.37$ & $54.54$ & $73.00$ & $75.15$ & $65.47$\\

\texttt{CIDER}            & $98.03$ & $48.64$ & $97.80$ & $41.06$ & $86.49$ & $42.86$ & $74.94$ & $62.46$ & $98.01$ & $48.64$ & $91.04$ & $43.09$ & $98.44$ & $39.78$ & $98.59$ & $38.10$\\
\texttt{SSD+}             & $99.10$ & $48.36$ & $99.84$ & $26.24$ & $99.47$ & $29.79$ & $95.86$ & $40.95$ & $94.99$ & $45.45$ & $99.38$ & $32.76$ & $85.60$ & $48.86$ & $98.62$ & $38.03$\\
\texttt{SNN}              & $22.60$ & $95.53$ & $\textbf{30.87}$ & $92.78$ & $25.14$ & $94.18$ & $29.87$ & $93.76$ & $30.74$ & $92.94$ & $34.26$ & $92.12$ & $31.43$ & $93.65$ & $\textbf{43.15}  $ &$87.13$ \\
\texttt{SCE}              & $19.71$ & $95.62$ & $34.53$ & $92.11$ & $22.87$ & $94.76$ & $\textbf{22.90}$ & $94.77$ & $24.42$ & $94.25$ & $24.81$ & $94.40$ & $31.97$ & $93.47$      & $46.13$ & $83.15$\\

 \texttt{GCE}              & $18.56$ & $96.33 $ & $35.75  $ & $91.50$ & $ 19.44   $ & $ 96.03 $ & $23.47$ & $95.11$ & $ 18.89  $ & $95.86$ & $19.78  $ & $95.61$ &$36.62  $ & $91.65$&$68.54$ & $77.54$     \\
\texttt{DivideMix}        & $40.81$ & $89.16$ & $39.32$ & $91.53$ & $65.83$ & $81.83$ & $66.64$ & $84.52$ & $59.27$ & $84.22$ & $24.81$ & $94.40$ &$34.27$    & $91.77 $ & $56.28$ & $82.92$\\

\texttt{Co-teaching}        & $81.94    $ & $ 58.15$ & $ 82.14  $ & $60.68$ & $77.98$ & $63.71$ & $53.42$ & $74.23$ & $77.59  $ & $62.76$ & $81.94  $ & $58.15$ &  $68.47$    & $61.72$ & $81.68$ & $59.61$\\

\texttt{CM}               & $18.32$ & $96.33$ & $36.28$ & $89.66$ & $\textbf{21.22}$ & $95.17$ & $24.72$ & $94.61$ & $23.62$ & $95.04$ & $\textbf{20.70}$ & $95.51$ & $33.50$ & $92.75$ & $49.52$ & $85.41$\\
\texttt{VolMinNet}        & $\textbf{15.00}$ & $96.89$ & $37.52$ & $91.60$ & $23.37$ & $94.96$ & $\textbf{22.90}$ & $95.15$ & $\textbf{18.80}$ & $96.15$ & $22.19$ & $95.02$ & $\textbf{29.26}$    & $94.09$   & $56.65$ & $81.39$\\
\midrule
\texttt{NOODLE}     & $\textbf{14.42}$ & $96.78$ & $\textbf{27.94}$ & $93.78$ & $\textbf{17.78}$ & $96.05$ & $\textbf{17.60}$ & $96.07$ & $\textbf{16.21}$ & $96.52$ & $\textbf{16.07}$ & $96.39$ & $\textbf{25.25}$ & $95.13$ & $\textbf{36.54}$ & $89.41$\\





\bottomrule
\end{tabular}
}
\end{table*}



 \paragraph{Implementation Settings.}
We use a CNN-based architecture, DenseNet-101\cite{densenet}, as the backbone model for all datasets. We train the model from scratch using the ID datasets. During training for CIFAR-10N and Animal-10N, we set the number of epochs to 100 and use a batch size of 64. First, we extract penultimate layer’s features and then apply  global average pooling following by $\ell_2$-normalization before performing the PI-based low-rank decomposition module of our NOODLE approach. We initialize the transition matrices as identity matrices of appropriate size in the case of CM-based approaches. For all datasets, we choose stochastic gradient descent (SGD) as the optimizer with a momentum of 0.9 and a weight decay of $1\times10^{-4}$. We tune the hyperparameters $\lambda$ from the set of values $\{0.0001, 0.0005, 0.001, 0.005, 0.1\}$. For the NOODLE approach, we consider different options for loss correction strategies such as CM and SCE. In terms of distance metrics in NOODLE approach, we consider both $k$NN and Mahalanobis scores as OOD detection metrics. We present the best performing variants of the NOODLE approach in the main result tables, yet present the detailed ablation study across different combinations of loss correction and distance metrics in the later sections.

\paragraph{Evaluation metrics.}
We evaluate the OOD detection performance using three widely recognized metrics. The false positive rate at 95\% true positive rate (FPR@95) indicates the proportion of OOD samples erroneously classified as ID when the true positive rate is fixed at 95\%; lower values correspond to better detection. The area under the receiver operating characteristic curve (AUROC) indicates the trade-off between true and false positive rates across thresholds and the higher value corresponds to better OOD performance. Finally, ID Accuracy (ID ACC) measures how accurately the model classifies the ID samples during testing. ID accuracy results are presented in the supplementary section.

\begin{figure*}[t]
    \centering
    \includegraphics[width=1\linewidth]{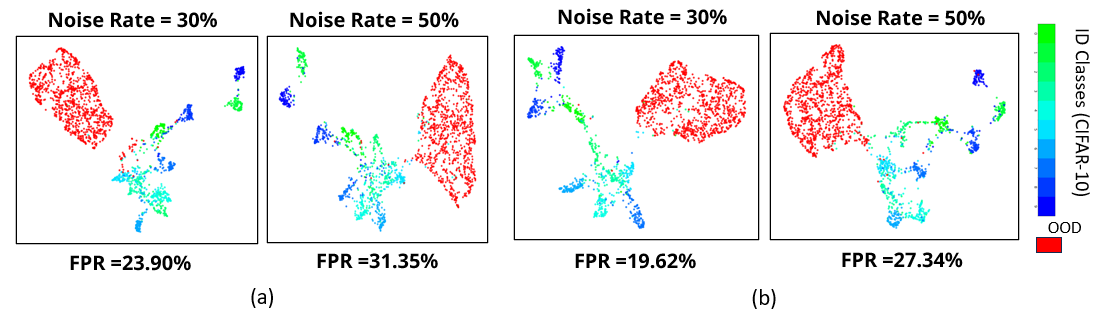}
   \caption{The effect of label noise on OOD detection for robust methods. The figure shows the UMAP representations of the latent feature vectors $\bm{h}(\bm{x})$ learned using (a) CM approach (transition matrix-based label noise correction) with cross-entropy loss, and (b) NOODLE, our proposed approach, on the CIFAR-10 dataset with synthetic label noise. The FPR95 metric for OOD detection is reported for each method under different label noise settings. While both methods mitigate the effect of noisy labels in learned features, NOODLE better preserves the ID-ness characteristics, reducing the mix-up of ID and OOD samples that results in improved OOD detection performance compared to CM.}

    \label{fig:enter-label}
\end{figure*}


\begin{figure}[t]
    \centering
    \begin{subfigure}[t]{0.48\linewidth}
        \centering
        \includegraphics[width=\linewidth]{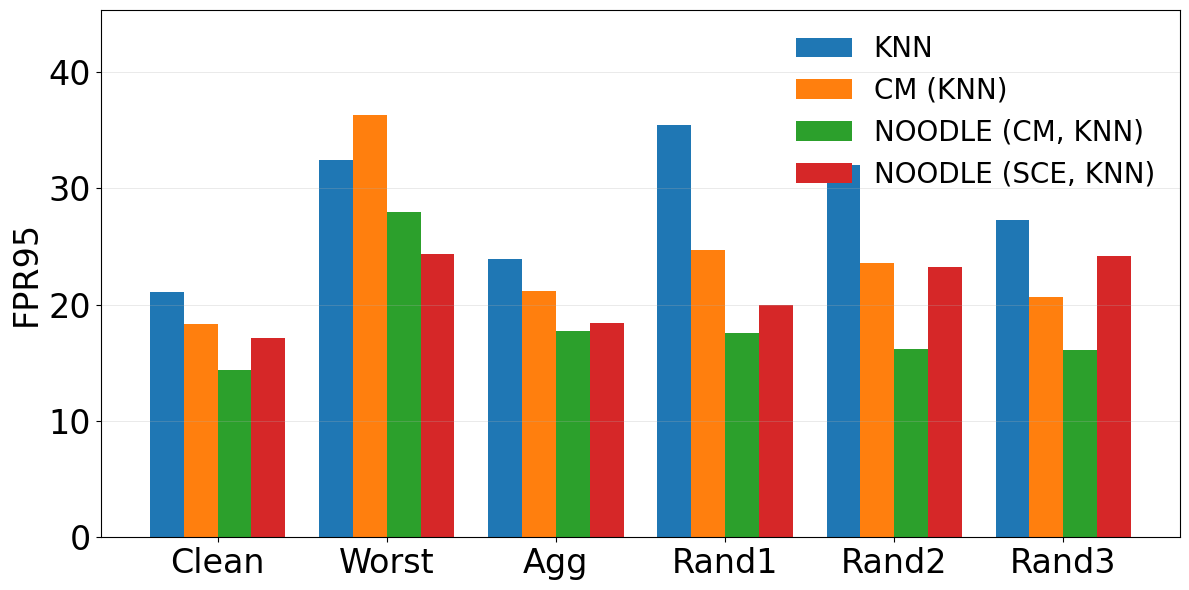}
        \caption{}
        \label{fig:cifar10n_knn_barplot}
    \end{subfigure}
        \hfill
    \begin{subfigure}[t]{0.48\linewidth}
        \centering
        \includegraphics[width=\linewidth]{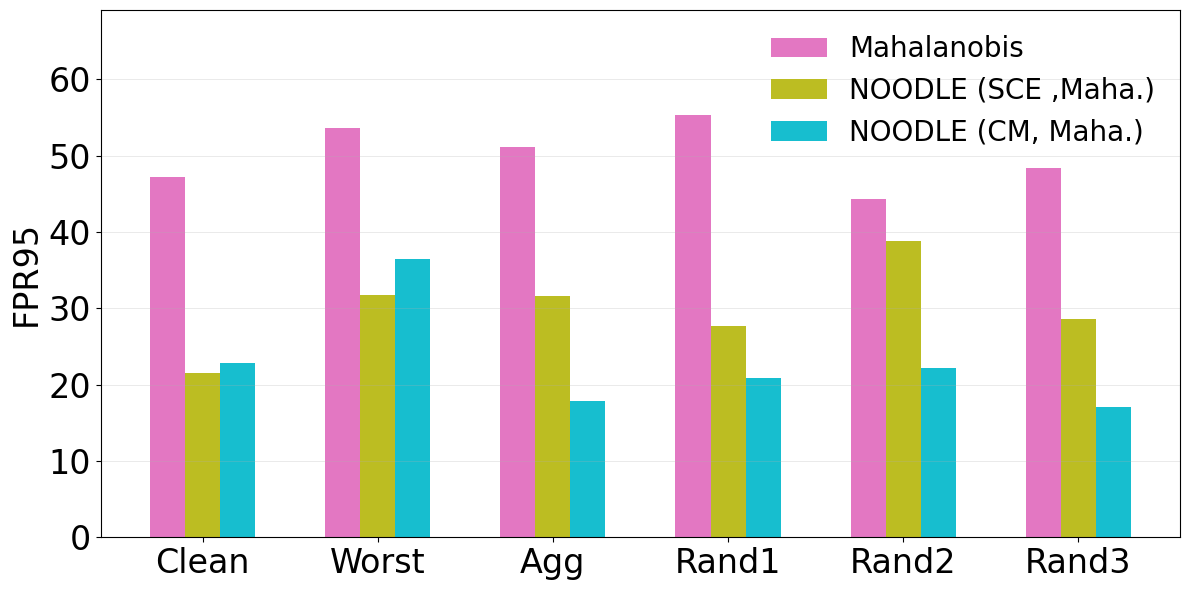}
        \caption{}
        \label{fig:cifar10n_maha_barplot}
    \end{subfigure}
    
    \caption{Comparison of OOD detection performance (FPR95$\downarrow$) on CIFAR-10N. 
    (a) Different KNN variants of NOODLE 
    (b) Different Mahalanobis variants of NOODLE. Here ``NOODLE (X, Y)" refers to NOODLE with X as loss correction strategy and Y as OOD distance metric. }
    \label{fig:cifar10n_barplots}
\end{figure}



\begin{figure}[t]
    \centering
    \begin{subfigure}[t]{0.3\linewidth}
        \centering
        \includegraphics[width=\linewidth]{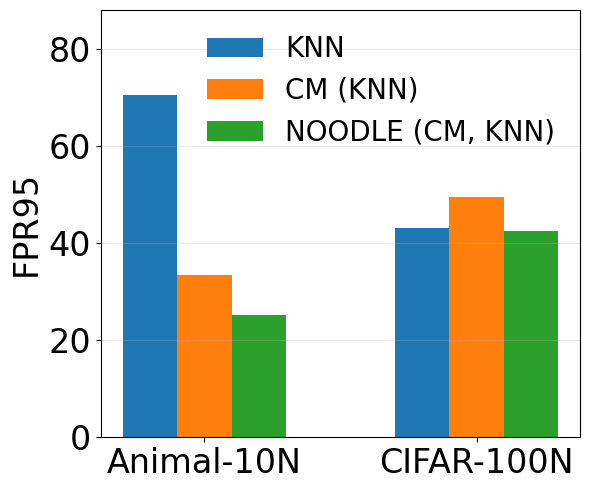}
        \caption{}
        \label{fig:cifar100n_barplot}
    \end{subfigure}
    \hfill
    \begin{subfigure}[t]{0.3\linewidth}
        \centering
        \includegraphics[width=\linewidth]{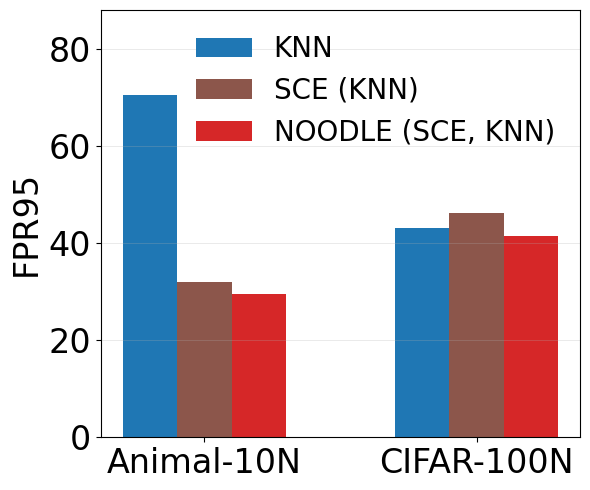}
        \caption{}
        \label{fig:animal10n_barplot}
    \end{subfigure}
    \hfill
    \begin{subfigure}[t]{0.3\linewidth}
        \centering
        \includegraphics[width=\linewidth]{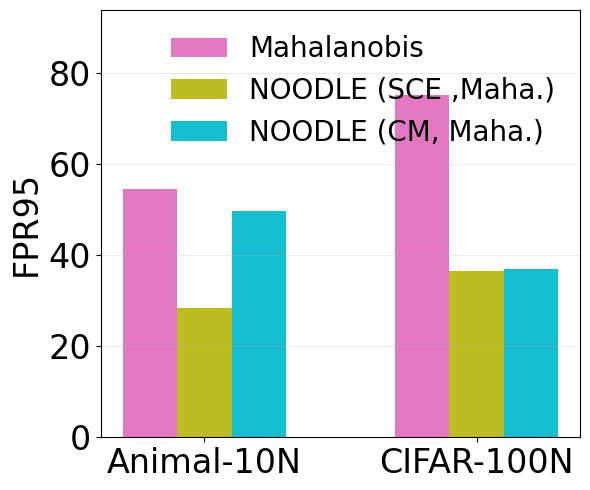}
        \caption{}
        \label{fig:animal10n_barplot}
    \end{subfigure}
    \caption{Comparison of OOD detection performance (FPR95$\downarrow$) on Animal-10N and CIFAR-100N datasets. 
(a) CM vs NOODLE with KNN metric (b) SCE vs NOODLE with KNN metric (c) Different Mahalanobis variants of NOODLE. Here ``NOODLE (X, Y)" refers to NOODLE with X as loss correction strategy and Y as OOD distance metric.}

    \label{fig:cifar10n_barplots}
\end{figure}

\paragraph{Results.}

Table~\ref{tab:cifar10n_symmetric} presents the OOD detection performance of the baselines and our method under symmetric label noise across different noise rates for CIFAR-10 dataset. We can observe that OOD detection baselines that lack label noise-robust training strategy are significantly impacted by high levels of label noise. In contrast, the label noise-robust approaches, especially those based on probabilistic modeling such as CM and VolMinNet maintain relatively strong performance under noisy conditions. Notably, our proposed method, {NOODLE}, consistently outperforms all other approaches under test in terms of both average FPR95 and AUROC. Our approach is particularly effective at higher noise rates. For example, at 50\% noise rate, NOODLE achieves the best performance, with an average FPR95 of 27.34\% which represents a reduction in FPR95 of up to 12.5\% compared to the best baseline method. 

{Table~\ref{tab:cifar10n-animal10n} presents the OOD detection performance on the real noise datasets which are annotated by unreliable crowd workers. For CIFAR-10N ``worst'' noise level (the noise rate is about 40.21\%), NOODLE achieves a {9.5\%} reduction in FPR95 and a {1.07\%} increase in AUROC compared to the best performing baseline SNN. Across other noise settings of CIFAR-10N as well, NOODLE consistently outperforms all baselines. A similar trend is observed on the Animal-10N dataset: while label noise-robust methods such as SCE, CM, and VolMinNet improve FPR95 over other non-robust techniques, NOODLE achieves an additional reduction of approximately 14\% compared to the closest baseline. For CIFAR-100N which is more challenging under noisy settings, NOODLE substantially outperforming all the baselines with an improvement of about 15\% in FPR95 compared to the best performing baseline. More experiment results and discussion are presented in the supplementary section.

}



\paragraph{ID and OOD Feature Representation.}
From Fig.~\ref{fig:noise_knn}, 
it is evident that higher noise levels distort the feature space, leading to less compact and more intermixed clusters. This feature distortion degrades the overall OOD detection performance, as we see in our experiments. To illustrate how label noise-robust methods mitigate this issue, we examine the UMAP visualizations in Fig.~\ref{fig:enter-label} where the learned features $\bm h(\bm x)$ of the test data for both ID and OOD samples are presented.
Here,  we compare the feature representations from one of the competing baseline, i.e., CM and our approach NOODLE.
 For CM, one can observe that cluster distortions are mitigated compared to the scenarios as in Fig.~\ref{fig:noise_knn}, showing that loss correction strategy helps in feature cleaning to some extend. Nonetheless, ID and OOD sample mixups are still present significantly, especially near the boundaries.

 In contrast, NOODLE produces more compact and well-separated clusters even under high noise rates with much reduced feature mix-up between ID and OOD samples.
  This implies that the low-rank sparse decomposition strategy in NOODLE is effective in better retaining the ID-ness characteristics of the learned features, which helps in separating the ID and OOD samples more effectively during testing. The tighter clusters in NOODLE’s feature space indicates that the samples from the same class are grouped better and the classes are kept more separate, which helps it achieving lower FPR95 than CM under the same noisy label conditions.

 \paragraph{Ablation Study}
Here, we study the effect of the low-rank and sparse decomposition module in the NOODLE framework under different loss correction strategies and OOD detection metrics. Specifically,  we analyze CM and SCE strategies for loss correction, and $k$NN and Mahalanobis scores for the feature distance-based OOD detection.
 Fig. \ref{fig:cifar10n_knn_barplot} shows how the NOODLE variant with CM as loss correction and $k$NN as the distance metric substantially advance the OOD detection performance for CIFAR-10N dataset. The SCE variant of the NOODLE version is also reasonably good, yet CM variant (i.e., NOODLE(CM, KNN)) performs much better in all scenarios in CIFAR-10N. 
For example, in worst case noise version, NOODLE (CM, KNN) reduces FPR95 to 27.94\% from 32.48\% by KNN, showing its robustness even in challenging settings. We can also observe similar improvement in performance in Fig. \ref{fig:cifar10n_maha_barplot}, where CM-Mahalanobis variant of the NOODLE also exhibits impressive OOD detection performance across scenarios. These results suggest that under different strategies of loss correction and various OOD detection metrics, the feature cleansing strategy of the NOODLE is effective in improving the ID-OOD separability.
We also present similar analysis for other datasets such as Animal-10N and CIFAR-100N in Fig. \ref{fig:cifar10n_barplots}.
In CIFAR-100N dataset, the SCE-Mahalanobis variant of NOODLE achieves the best performance, likely because estimating transition matrices for CM-based methods becomes increasingly difficult as the number of classes grows. Nevertheless, our key idea of feature cleaning via low-rank sparse decomposition consistently enhances performance across different settings.

\section{Conclusion}
In this work, we introduce a novel framework for OOD detection under noisy labels that addresses the limitations of existing methods by correcting label noise and enhancing OOD performance. By leveraging low-rank ID feature representations and a carefully designed learning criterion, our approach provides greater flexibility and effectiveness in improving ID–OOD separability, even in highly noisy settings. Experimental results across multiple benchmarks and challenging OOD scenarios demonstrate the superiority of our method, highlighting its ability to tackle the challenging problem of OOD detection under noisy labels.

{
    \bibliographystyle{unsrt}
    \bibliography{neurips2025}
}


\clearpage

\begin{center}
    {\bf Supplementary Material of ``Tackling the Noisy Elephant in the Room: Label Noise-robust Out-of-Distribution Detection via Loss Correction and Low-rank Decomposition''}
\end{center}
\appendix


\section{Notation}\label{app:notation}
We use the following notation throughout the paper: $x$, $\x$, $\X$, and $\tX$ represent a scalar, a vector, a matrix, and a tensor, respectively. Both $x_i$ and $[\bm x]_i$ denote the $i$th entry of the vector $\x$.  $[\bm X]_{i,j}$ denote the $(i,j)$th entry of the matrix $\bm X$. $\bm x_i$ denotes the $i$th row of the matrix $\bm X$;
$[I]$ means an integer set $\{1,2,\ldots,I\}$. 
  $^\T$  denote transpose.
  $\bm X \ge \bm 0$ implies that all the entries of the matrix $\bm X$ are non-negative. $\mathbb{I}[A]$ denotes an indicator function for the event $A$ such that $\mathbb{I}[A]=1$ if the event $A$ happens, otherwise $\mathbb{I}[A]=0$. ${\sf CE}(\bm x,y) = -\sum_{k=1}^K \mathbb{I}[{y}=k]\log(\bm x(k))$ denotes the cross entropy function. 
  $\bm I$ denotes an identity matrix of appropriate size. $\bm 1_K$ denotes an all-one vector of size $K$. $|{\cal C}|$ denotes the cardinality of the set ${\cal C}.$ $\Delta^{K}$ denotes a $(K-1)$-dimensional probability simplex such that
     $\Delta^{K} = \{\bm u \in \mathbb{R}^K~| \bm u \ge \bm 0, \bm 1^{\top}\bm u=1\}$.

\section{Algorithm Description}

In this section, we present the NOODLE algorithm. 
Algorithm~\ref{alg:proposed-algo} provides the complete, step-by-step procedure of our approach using the transition matrix-based loss correction strategy. 
As discussed earlier, we obtain ID features via a low-rank sparse decomposition. 
The decomposition routine is detailed in Algorithm~\ref{alg:power-iteration}.

\begin{algorithm}
\caption{Proposed approach NOODLE }

\begin{algorithmic}[1]
\label{alg:proposed-algo}
\item[]  \textbf{Input:} Noisily labeled data \{$(\bm{x}_n,\widehat{y}_n)\}_{n=1}^N $, where $\bm{x}_n \in \mathcal{X}$, $\widehat{y}_n \in \mathcal{Y}$, $n_{\mathrm{iter}}$, stopping criterion,$K$ as number of classes
\item[] \textbf{Output:} Estimated parameters $\bm{\theta}$ and $\bm{T}$

\STATE Initialize Transition Matrix $ \bm T$ to identity matrices $ \bm I_K$ 

\STATE Initialize the parameters $\bm \theta$ of the neural network function class $ \mathcal{F}$

\WHILE{stopping criterion is not reached}
    \WHILE{stopping criterion is not reached}
        \STATE Draw a random batch $\mathcal{B}$
        \STATE $\bm{H} \gets [\,\bm{h}(\bm{x}_1), \dots, \bm{h}(\bm{x}_N)\,]$ \hfill\textit{// features from batch $\mathcal{B}$ as per Eq. \ref{eq:feature-matrix}}
        \STATE $\bm Q  \leftarrow $ ApproxTopKSingularVectors($\bm{H}, K, n_{\mathrm{iter}}$)
        \STATE  $\bm H_{\text{ID}} \leftarrow (\bm Q \bm Q^\top) \bm H$
         \STATE  $\bm H_{\text{OOD}} \leftarrow \bm H - \bm H_{\text{ID}}$

        \STATE Compute $\nabla \mathcal{L}_F(\bm{T}, \mathcal{B},\bm H_{OOD}))$
        \STATE $\bm{T}, \bm \theta \leftarrow \mathrm{SGDOptimizer}(\bm{T}, \nabla \mathcal{L}_F(\bm{T}, \mathcal{B},\bm H_{\text{OOD}}))$
    \ENDWHILE
\ENDWHILE
\end{algorithmic}
\end{algorithm}

\begin{algorithm}[t]
\caption{ApproxTopKSingularVectors }
\label{alg:power-iteration}
\begin{algorithmic}[1]
\item[] \textbf{Input:} Feature matrix $\bm{H} \in \mathbb{R}^{N \times D}$, target rank $k$, number of iterations $n_{\mathrm{iter}}$
\item[]\textbf{Output:} Orthonormal matrix $\bm{Q} \in \mathbb{R}^{D \times k}$ spanning the approximate top-$k$ right singular vectors of $\bm{H}$
\STATE Randomly initialize $\bm{Q} \in \mathbb{R}^{D \times k}$ \hfill\textit{// $D$: feature dimension, $k$: target rank}
\FOR{$i = 1$ to $n_{\mathrm{iter}}$}
    \STATE $\bm{Z} \gets \bm{H}^\top (\bm{H}\bm{Q})$ \hfill\textit{// project $\bm{Q}$ into column space of $\bm{H}$}
     \STATE $\bm{Q} \gets \mathrm{QRDecomposition}(\bm{Z})$ \hfill\textit{// obtain orthonormal basis of $\bm{Z}$’s column space}
\ENDFOR
\STATE \textbf{return} $\bm{Q}$ \hfill\textit{// spans approximate top-$k$ right singular vectors of $\bm{H}$}
\end{algorithmic}
\end{algorithm}

\section{More Experiment Results}
In this section, we present more detailed evaluations. While the summary results for CIFAR-10N were reported earlier, we now provide dataset-wise OOD performance along with ID accuracy in Table~\ref{tab:cifar-10-detail-1} and Table~\ref{tab:cifar-10-detail-2}. To ensure fairness, all post-hoc methods are evaluated using the same encoder trained with cross-entropy loss, thereby avoiding any bias in performance analysis. For CIDER and SSD+, we follow prior work but replace their default ResNet-18 encoder with DenseNet-101 for consistency. As a result, these methods may require additional fine-tuning to fully realize their potential. For Animal-10N, the dataset-specific results are reported in Table~\ref{tab:animal10n-detail}. We find that most baseline methods struggle to achieve a good balance between ID accuracy and OOD detection. In contrast, our proposed method NOODLE delivers consistently strong results across both metrics.  

Finally, detailed results on CIFAR-100 are shown in Table~\ref{tab:cifar100-detail}. As expected, CIFAR-100 is considerably more challenging, leading to significant performance degradation for most baselines. Nevertheless, NOODLE achieves the best OOD detection performance while maintaining a competitive and balanced ID accuracy, highlighting its robustness under difficult conditions.

\begin{table*}[t]
\centering
\caption{OOD detection performance (FPR95$\downarrow$ / AUROC$\uparrow$) on CIFAR-10 under different noise settings using a DenseNet-100 encoder.}
\label{tab:cifar-10-detail-1}
\setlength{\tabcolsep}{4pt}
\renewcommand{\arraystretch}{1.2}
\resizebox{\textwidth}{!}{
\begin{tabular}{lccccccccccccccc}
\toprule
\textbf{Method}
& \multicolumn{2}{c}{SVHN}
& \multicolumn{2}{c}{FashionMNIST}
& \multicolumn{2}{c}{LSUN}
& \multicolumn{2}{c}{iSUN}
& \multicolumn{2}{c}{Texture}
& \multicolumn{2}{c}{Places365}
& \multicolumn{2}{c}{Average}
& \textbf{ID Acc.} \\
\cmidrule(lr){2-3}\cmidrule(lr){4-5}\cmidrule(lr){6-7}\cmidrule(lr){8-9}\cmidrule(lr){10-11}\cmidrule(lr){12-13}\cmidrule(lr){14-15}\cmidrule(lr){16-16}
& FPR95$\downarrow$ & AUROC$\uparrow$
& FPR95$\downarrow$ & AUROC$\uparrow$
& FPR95$\downarrow$ & AUROC$\uparrow$
& FPR95$\downarrow$ & AUROC$\uparrow$
& FPR95$\downarrow$ & AUROC$\uparrow$
& FPR95$\downarrow$ & AUROC$\uparrow$
& FPR95$\downarrow$ & AUROC$\uparrow$
& \\
\midrule
\multicolumn{16}{c}{\textbf{Clean}} \\
\midrule
\texttt{KNN}         & $10.25$ & $98.26$ & $10.95$ & $98.03$ & $13.21$ & $97.64$ & $17.05$ & $96.87$ & $25.30$ & $95.31$ & $49.58$ & $88.70$ & $21.06$ & $95.80$ & $93.32$  \\
\texttt{MSP}         & $72.54$ & $87.37$ & $49.86$ & $92.74$ & $34.70$ & $95.33$ & $46.04$ & $93.34$ & $68.51$ & $85.16$ & $66.90$ & $86.46$ & $56.43$ & $90.07$ & $93.32$  \\
\texttt{ODIN}        & $55.88$ & $89.16$ & $16.01$ & $97.23$ & $3.01$ & $99.12$ & $8.47$ & $98.20$ & $60.46$ & $82.84$ & $54.77$ & $88.30$ & $33.10$ & $92.47$ & $93.32$  \\
\texttt{Energy}      & $73.05$ & $87.23$ & $15.64$ & $97.18$ & $4.44$ & $98.86$ & $23.22$ & $96.16$ & $67.02$ & $83.48$ & $51.52$ & $89.30$ & $39.15$ & $92.03$ & $93.32$  \\
\texttt{ReAct}       & $97.03$ & $61.00$ & $44.59$ & $93.52$ & $28.01$ & $95.84$ & $41.26$ & $93.42$ & $88.14$ & $68.86$ & $62.85$ & $86.07$ & $60.31$ & $83.12$ & $93.32$  \\
\texttt{Mahalanobis} & $4.51$  & $99.13$ & $2.47$  & $99.31$ & $0.63$  & $99.75$ & $14.79$ & $97.38$ & $22.46$ & $95.08$ & $69.08$ & $82.08$ & $18.99$ & $95.45$ & $93.32$  \\
\texttt{CIDER}       & $89.25$ & $86.21$ & $100.00$ & $46.56$ & $100.00$ & $51.96$ & $100.00$ & $29.81$ & $99.04$ & $35.07$ & $99.90$ & $42.25$ & $98.03$ & $48.64$ & $94.03$ \\
\texttt{SSD+}        & $99.25$ & $62.41$ & $96.42$ & $53.98$ & $100.00$ & $42.95$ & $99.90$ & $43.05$ & $99.18$ & $41.64$ & $99.87$ & $46.14$ & $99.10$ & $48.36$ & $94.03$ \\
\texttt{SNN}         & $8.68$  & $98.35$ & $21.49$ & $96.22$ & $9.22$  & $98.42$ & $19.46$ & $96.72$ & $26.99$ & $94.97$ & $49.74$ & $88.52$ & $22.60$ & $95.53$ & $94.15$ \\
\texttt{SCE}         & $4.59$  & $99.13$ & $15.47$ & $97.06$ & $1.96$  & $99.58$ & $10.60$ & $98.06$ & $29.31$ & $92.84$ & $56.36$ & $87.03$ & $19.71$ & $95.62$ & $91.09$ \\
\texttt{GCE}         & $11.33$ & $98.02$ & $11.73$ & $98.00$ & $7.16$  & $98.73$ & $9.06$  & $98.30$ & $21.70$ & $96.15$ & $50.39$ & $88.80$ & $18.56$ & $96.33$ & $93.54$ \\

\texttt{DivideMix} 
& $22.35$ & $94.72$ 
& $36.90$ & $91.74$ 
& $33.41$ & $90.87$ 
& $46.05$ & $87.62$ 
& $47.11$ & $86.55$ 
& $58.04$ & $84.45$ 
& $40.81$ & $89.16$ 
& $81.22$ \\

\texttt{Co-teaching} & $56.55$ & $74.88$ & $99.90$ & $36.86$ & $95.18$ & $58.31$ & $96.28$ & $61.89$ & $55.83$ & $78.99$ & $94.03$ & $48.08$ & $81.94$ & $58.15$ & $86.22$ \\
\texttt{CM}          & $8.35$  & $98.64$ & $8.39$  & $98.49$ & $7.16$  & $98.67$ & $15.19$ & $97.31$ & $19.72$ & $96.36$ & $51.14$ & $88.54$ & $18.32$ & $96.33$ & $94.39$ \\
\texttt{VolMinNet}   & $2.45$  & $99.48$ & $13.06$ & $97.57$ & $5.96$  & $98.93$ & $6.25$  & $98.85$ & $15.96$ & $97.07$ & $46.30$ & $89.42$ & $15.00$ & $96.89$ & $94.56$ \\
\midrule
\texttt{NOODLE}      & $2.78$  & $99.39$ & $5.05$  & $98.88$ & $4.79$  & $99.07$ & $10.05$ & $97.92$ & $15.85$ & $96.80$ & $48.02$ & $88.61$ & $14.42$ & $96.78$ & $94.29$ \\
\midrule
\multicolumn{16}{c}{\textbf{Noise = Worst}} \\
\midrule
\texttt{KNN}         & $9.17$  & $98.28$ & $27.89$ & $94.95$ & $15.76$ & $96.98$ & $38.41$ & $92.03$ & $36.21$ & $90.94$ & $67.46$ & $84.15$ & $32.48$ & $92.89$ & $80.79$ \\
\texttt{MSP}         & $56.74$ & $87.97$ & $50.54$ & $89.33$ & $38.83$ & $93.72$ & $62.58$ & $87.03$ & $78.16$ & $76.44$ & $74.04$ & $78.44$ & $60.15$ & $85.49$ & $80.79$ \\
\texttt{ODIN}        & $45.45$ & $90.96$ & $33.09$ & $93.31$ & $13.41$ & $97.43$ & $32.36$ & $93.02$ & $71.68$ & $72.39$ & $75.89$ & $73.02$ & $45.31$ & $86.69$ & $80.79$ \\
\texttt{Energy}      & $37.62$ & $93.17$ & $35.78$ & $93.28$ & $14.57$ & $97.00$ & $47.39$ & $90.98$ & $76.06$ & $74.61$ & $72.39$ & $77.45$ & $47.30$ & $87.75$ &$80.79$ \\
\texttt{ReAct}       & $70.94$ & $82.60$ & $52.84$ & $90.05$ & $36.40$ & $93.87$ & $68.31$ & $81.53$ & $88.92$ & $54.21$ & $77.90$ & $70.87$ & $65.89$ & $78.86$ & $80.79$ \\
\texttt{Mahalanobis} & $4.41$  & $98.97$ & $19.95$ & $96.33$ & $10.05$ & $98.07$ & $30.50$ & $92.65$ & $33.69$ & $89.11$ & $87.96$ & $60.38$ & $31.09$ & $89.25$ & $80.79$ \\
\texttt{CIDER}       & $99.88$ & $30.48$ & $96.13$ & $49.01$ & $99.94$ & $33.75$ & $99.21$ & $35.94$ & $92.91$ & $54.56$ & $98.73$ & $42.62$ & $97.80$ & $41.06$ & $24.22$ \\
\texttt{SSD+}        & $99.96$ & $43.94$ & $99.94$ & $33.16$ & $99.98$ & $13.02$ & $100.00$ & $16.11$ & $99.50$ & $18.12$ & $99.67$ & $33.06$ & $99.84$ & $26.24$ & $19.85$ \\
\texttt{SNN}         & $10.98$ & $97.95$ & $24.49$ & $95.67$ & $16.61$ & $96.76$ & $31.84$ & $93.96$ & $33.92$ & $90.69$ & $67.36$ & $81.67$ & $30.87$ & $92.78$ & $80.57$\\
\texttt{SCE}         & $9.84$  & $98.12$ & $19.48$ & $96.06$ & $12.17$ & $97.41$ & $12.14$ & $97.41$ & $35.11$ & $91.21$ & $64.89$ & $82.63$ & $25.61$ & $93.81$ & $83.48$ \\
\texttt{GCE}         & $10.53$ & $98.18$ & $21.72$ & $96.09$ & $15.71$ & $97.05$ & $55.81$ & $87.25$ & $45.69$ & $87.86$ & $65.02$ & $82.54$ & $35.75$ & $91.50$ & $83.49$ \\
\texttt{DivideMix} 
& $44.70$ & $93.52$ 
& $48.20$ & $89.20$ 
& $38.09$ & $90.29$ 
& $15.35$ & $96.70$ 
& $32.06$ & $94.42$ 
& $57.95$ & $85.09$ 
& $39.39$ & $91.53$ 
& $80.23$ \\

\texttt{Co-teaching} & $50.04$ & $85.37$ & $99.95$ & $29.04$ & $99.61$ & $58.63$ & $96.88$ & $58.36$ & $52.15$ & $82.65$ & $94.22$ & $50.03$ & $82.14$ & $60.68$ & $86.34$ \\
\texttt{CM}          & $8.61$  & $98.29$ & $17.92$ & $96.78$ & $28.37$ & $90.87$ & $56.29$ & $81.97$ & $39.24$ & $88.19$ & $67.23$ & $81.87$ & $36.28$ & $89.66$ & $76.26$ \\
\texttt{VolMinNet}   & $7.58$  & $98.33$ & $42.88$ & $90.42$ & $19.70$ & $95.50$ & $43.71$ & $89.08$ & $40.90$ & $87.46$ & $70.64$ & $77.54$ & $37.57$ & $89.72$ & $76.30$ \\
\midrule
\texttt{NOODLE}      & $3.23$  & $99.31$ & $23.67$ & $95.81$ & $8.08$  & $98.40$ & $36.55$ & $93.57$ & $32.98$ & $91.67$ & $63.14$ & $83.91$ & $27.94$ & $93.78$ & $83.54$ \\
\midrule
\multicolumn{16}{c}{\textbf{Noise = Aggre}} \\
\midrule
\texttt{KNN}         & $11.37$ & $97.91$ & $18.32$ & $96.70$ & $9.93$ & $98.11$ & $18.12$ & $96.35$ & $23.69$ & $95.43$ & $62.26$ & $84.54$ & $23.95$ & $94.84$ & $89.93$ \\
\texttt{MSP}         & $57.75$ & $89.45$ & $47.16$ & $90.89$ & $27.74$ & $95.11$ & $59.89$ & $88.27$ & $67.84$ & $83.15$ & $69.89$ & $82.42$ & $55.04$ & $88.21$ & $89.93$ \\
\texttt{ODIN}        & $49.75$ & $91.54$ & $30.78$ & $94.69$ & $9.29$ & $98.25$ & $33.50$ & $92.13$ & $65.18$ & $82.01$ & $74.50$ & $75.54$ & $43.83$ & $89.03$ & $89.93$ \\
\texttt{Energy}      & $66.66$ & $88.00$ & $40.23$ & $93.66$ & $18.07$ & $97.10$ & $64.95$ & $88.41$ & $71.46$ & $81.96$ & $74.80$ & $78.26$ & $56.03$ & $87.90$ & $89.93$ \\
\texttt{ReAct}       & $66.66$ & $88.00$ & $40.23$ & $93.66$ & $18.07$ & $97.10$ & $64.95$ & $88.41$ & $71.46$ & $81.96$ & $21.61$ & $96.92$ & $47.17$ & $91.01$ & $89.93$ \\
\texttt{Mahalanobis} & $4.41$  & $98.97$ & $19.95$ & $96.33$ & $10.05$ & $98.07$ & $30.50$ & $92.65$ & $33.69$ & $89.11$ & $87.96$ & $60.38$ & $31.09$ & $89.25$ & $89.93$ \\
\texttt{CIDER}       & $28.18$ & $91.29$ & $99.98$ & $34.71$ & $96.13$ & $49.01$ & $100.00$ & $22.14$ & $94.47$ & $34.71$ & $96.37$ & $48.95$ & $86.49$ & $42.86$ & $25.24$ \\
\texttt{SSD+}        & $99.64$ & $43.12$ & $99.90$ & $31.72$ & $99.87$ & $36.56$ & $100.00$ & $21.46$ & $97.41$ & $39.38$ & $100.00$ & $6.48$  & $99.47$ & $29.79$ & $19.22$\\
\texttt{SNN}         & $6.15$  & $98.90$ & $14.36$ & $97.33$ & $8.61$  & $98.32$ & $35.76$ & $92.37$ & $26.12$ & $94.45$ & $59.86$ & $83.74$ & $25.14$ & $94.18$ & $90.00$ \\
\texttt{SCE}         & $4.32$  & $99.14$ & $19.12$ & $96.41$ & $4.92$  & $98.98$ & $19.90$ & $96.39$ & $24.98$ & $94.82$ & $63.95$ & $82.82$ & $22.87$ & $94.76$ & $90.45$ \\
\texttt{GCE}         & $1.94$  & $99.64$ & $11.17$ & $98.11$ & $7.57$  & $98.58$ & $14.44$ & $97.36$ & $28.63$ & $94.62$ & $52.86$ & $87.84$ & $19.43$ & $96.03$ & $91.55$ \\

\texttt{DivideMix} 
& $77.88$ & $85.56$ 
& $84.07$ & $74.54$ 
& $48.41$ & $87.26$ 
& $59.81$ & $78.07$ 
& $49.49$ & $86.58$ 
& $74.63$ & $78.99$ 
& $65.72$ & $81.83$ 
& $ 73.44$ \\

\texttt{Co-teaching} & $48.52$ & $80.16$ & $99.49$ & $39.54$ & $94.18$ & $58.15$ & $86.59$ & $72.57$ & $45.46$ & $84.21$ & $93.64$ & $47.65$ & $77.98$ & $63.71$ & $86.15$ \\
\texttt{CM}          & $7.82$  & $98.68$ & $13.01$ & $97.55$ & $8.41$  & $98.30$ & $12.70$ & $97.62$ & $22.45$ & $95.18$ & $62.91$ & $83.70$ & $21.22$ & $95.17$ & $78.82$ \\
\texttt{VolMinNet}   & $3.88$  & $99.17$ & $10.28$ & $98.08$ & $8.46$  & $98.32$ & $31.84$ & $93.84$ & $29.41$ & $93.06$ & $56.36$ & $87.27$ & $23.37$ & $94.96$ & $91.86$ \\
\midrule
\texttt{NOODLE}      & $1.44$  & $99.67$ & $16.17$ & $97.17$ & $3.42$  & $99.24$ & $9.69$  & $98.15$ & $19.26$ & $95.90$ & $56.68$ & $86.19$ & $17.78$ & $96.05$ & $91.57$ \\
\bottomrule
\end{tabular}
}
\end{table*}

\begin{table*}[t]
\centering
\caption{OOD detection performance (FPR95$\downarrow$ / AUROC$\uparrow$) on CIFAR-10 using a DenseNet-100 encoder under \textbf{random1}, \textbf{random2}, and \textbf{random3} noise.}
\label{tab:cifar-10-detail-2}
\setlength{\tabcolsep}{4pt}
\renewcommand{\arraystretch}{1.2}
\resizebox{\textwidth}{!}{
\begin{tabular}{lccccccccccccccc}
\toprule
\textbf{Method}
& \multicolumn{2}{c}{SVHN}
& \multicolumn{2}{c}{FashionMNIST}
& \multicolumn{2}{c}{LSUN}
& \multicolumn{2}{c}{iSUN}
& \multicolumn{2}{c}{Texture}
& \multicolumn{2}{c}{Places365}
& \multicolumn{2}{c}{Average}
& \textbf{ID Acc.} \\
\cmidrule(lr){2-3}\cmidrule(lr){4-5}\cmidrule(lr){6-7}\cmidrule(lr){8-9}\cmidrule(lr){10-11}\cmidrule(lr){12-13}\cmidrule(lr){14-15}\cmidrule(lr){16-16}
& FPR95$\downarrow$ & AUROC$\uparrow$
& FPR95$\downarrow$ & AUROC$\uparrow$
& FPR95$\downarrow$ & AUROC$\uparrow$
& FPR95$\downarrow$ & AUROC$\uparrow$
& FPR95$\downarrow$ & AUROC$\uparrow$
& FPR95$\downarrow$ & AUROC$\uparrow$
& FPR95$\downarrow$ & AUROC$\uparrow$
& \\
\midrule
\multicolumn{16}{c}{\textbf{Noise = random1}} \\
\midrule
\texttt{KNN}
& $26.55$ & $95.11$ & $32.22$ & $94.66$ & $19.98$ & $96.43$ & $36.27$ & $92.58$
& $30.12$ & $93.01$ & $67.75$ & $84.12$ & $35.48$ & $92.65$ & $88.03$ \\
\texttt{MSP}
& $63.91$ & $89.44$ & $56.80$ & $88.35$ & $33.35$ & $94.64$ & $70.25$ & $83.87$
& $70.71$ & $81.70$ & $69.47$ & $81.73$ & $60.75$ & $86.62$ & $88.03$ \\
\texttt{ODIN}
& $55.83$ & $87.93$ & $43.60$ & $91.41$ & $11.17$ & $98.01$ & $46.12$ & $89.32$
& $65.07$ & $77.53$ & $72.94$ & $75.16$ & $49.12$ & $86.56$ & $88.03$ \\
\texttt{Energy}
& $81.76$ & $85.45$ & $39.75$ & $92.56$ & $11.32$ & $97.78$ & $63.37$ & $86.34$
& $68.55$ & $80.04$ & $64.57$ & $81.25$ & $54.89$ & $87.24$ & $88.03$ \\
\texttt{ReAct}
& $97.67$ & $60.00$ & $57.46$ & $89.81$ & $28.41$ & $95.32$ & $72.90$ & $78.42$
& $86.65$ & $62.40$ & $69.53$ & $75.58$ & $68.77$ & $76.92$ & $88.03$\\
\texttt{Mahalanobis}
& $16.86$ & $96.22$ & $48.89$ & $88.49$ & $13.62$ & $97.36$ & $50.63$ & $82.90$
& $36.24$ & $87.86$ & $92.57$ & $50.57$ & $43.14$ & $83.90$ & $88.03$ \\
\texttt{CIDER}
& $54.22$ & $82.56$ & $69.96$ & $72.92$ & $96.13$ & $49.01$ & $100.00$ & $22.14$
& $94.47$ & $34.71$ & $93.35$ & $48.34$ & $74.94$ & $62.46$ & $26.02$ \\
\texttt{SSD+}
& $90.68$ & $45.50$ & $94.43$ & $48.88$ & $99.66$ & $29.40$ & $99.66$ & $35.47$
& $92.61$ & $39.44$ & $98.11$ & $47.00$ & $95.86$ & $40.95$ & $22.05$\\
\texttt{SNN}
& $16.25$ & $97.07$ & $35.63$ & $93.93$ & $15.74$ & $97.05$ & $23.81$ & $95.32$
& $26.77$ & $93.92$ & $61.03$ & $85.28$ & $29.87$ & $93.76$ & $87.13$\\
\texttt{SCE}
& $17.45$ & $97.24$ & $10.51$ & $97.91$ & $5.81$ & $98.81$ & $14.95$ & $97.11$
& $25.83$ & $93.70$ & $62.82$ & $83.86$ & $22.90$ & $94.77$ & $89.81$ \\
\texttt{GCE}
& $6.38$ & $98.87$ & $11.85$ & $97.92$ & $11.48$ & $97.93$ & $23.18$ & $96.11$
& $31.91$ & $93.27$ & $56.04$ & $86.58$ & $23.47$ & $95.11$ & $90.46$ \\
\texttt{DivideMix} 
& $89.21$ & $84.82$ 
& $62.24$ & $86.97$ 
& $56.76$ & $88.52$ 
& $68.00$ & $80.74$ 
& $50.85$ & $86.88$ 
& $72.78$ & $79.19$ 
& $66.64$ & $84.52$ 
& $77.14$ \\

\texttt{Co-teaching}
& $51.32$ & $81.34$ & $49.00$ & $83.00$ & $52.00$ & $87.00$ & $23.18$ & $96.11$
& $74.58$ & $69.89$ & $94.74$ & $62.07$ & $53.42$ & $74.23$ & $86.44$ \\
\texttt{CM}
& $3.13$ & $99.35$ & $18.50$ & $96.55$ & $15.74$ & $97.22$ & $21.04$ & $96.18$
& $25.85$ & $94.65$ & $64.08$ & $83.69$ & $24.72$ & $94.61$ & $90.31$\\
\texttt{VolMinNet}
& $5.48$ & $99.01$ & $15.74$ & $97.11$ & $6.92$ & $98.75$ & $22.86$ & $95.86$
& $30.30$ & $94.07$ & $56.10$ & $86.06$ & $22.90$ & $95.15$ & $90.53$\\
\midrule

\texttt{NOODLE}
& $4.04$ & $99.21$ & $17.33$ & $96.70$ & $4.15$ & $99.18$ & $6.37$ & $98.65$
& $14.57$ & $97.11$ & $59.12$ & $85.57$ & $17.60$ & $96.07$ & $90.52$ \\
\midrule
\multicolumn{16}{c}{\textbf{Noise = random2}} \\
\midrule
\texttt{KNN}
& $6.29$ & $98.73$ & $41.05$ & $91.75$ & $19.28$ & $96.04$ & $24.92$ & $95.02$
& $30.85$ & $92.20$ & $69.53$ & $82.43$ & $31.99$ & $92.70$ & $87.79$ \\
\texttt{MSP}
& $53.36$ & $89.13$ & $58.24$ & $84.86$ & $30.82$ & $94.26$ & $48.26$ & $91.41$
& $73.00$ & $80.23$ & $74.98$ & $79.10$ & $56.44$ & $86.50$ & $87.79$ \\
\texttt{ODIN}
& $44.99$ & $87.97$ & $56.14$ & $83.96$ & $13.74$ & $97.24$ & $19.78$ & $94.78$
& $68.03$ & $73.23$ & $77.58$ & $68.27$ & $46.71$ & $84.24$ & $87.79$\\
\texttt{Energy}
& $60.00$ & $87.44$ & $50.75$ & $87.08$ & $17.62$ & $96.63$ & $29.00$ & $93.97$
& $71.31$ & $76.10$ & $73.20$ & $75.42$ & $50.31$ & $86.11$ & $87.79$ \\
\texttt{ReAct}
& $78.12$ & $80.74$ & $56.58$ & $86.77$ & $30.49$ & $94.12$ & $62.46$ & $86.31$
& $87.27$ & $61.33$ & $77.87$ & $69.69$ & $65.46$ & $79.83$ & $87.79$ \\
\texttt{Mahalanobis}
& $11.94$ & $96.96$ & $69.11$ & $81.20$ & $23.57$ & $95.12$ & $36.79$ & $87.81$
& $38.72$ & $86.17$ & $94.13$ & $48.44$ & $45.71$ & $82.62$ & $87.79$\\
\texttt{CIDER}
& $89.16$ & $86.14$ & $99.99$ & $46.58$ & $99.90$ & $42.25$ & $100.00$ & $29.81$
& $99.04$ & $35.07$ & $100.00$ & $51.96$ & $98.01$ & $48.64$ & $25.46$ \\
\texttt{SSD+}
& $99.36$ & $35.10$ & $90.31$ & $57.22$ & $93.77$ & $55.07$ & $99.56$ & $33.89$
& $91.78$ & $42.09$ & $95.15$ & $49.31$ & $94.99$ & $45.45$ & $23.45$ \\
\texttt{SNN}
& $4.84$ & $99.05$ & $42.45$ & $92.36$ & $19.44$ & $96.04$ & $17.68$ & $96.06$
& $33.90$ & $91.82$ & $66.09$ & $82.28$ & $30.74$ & $92.94$ & $88.11$ \\
\texttt{SCE}
& $6.17$ & $98.78$ & $15.97$ & $97.16$ & $13.15$ & $97.50$ & $24.60$ & $95.04$
& $25.41$ & $94.60$ & $61.19$ & $82.42$ & $24.42$ & $94.25$ & $89.72$ \\
\texttt{GCE}
& $2.02$ & $99.57$ & $18.91$ & $96.46$ & $5.57$ & $98.84$ & $9.06$ & $98.09$
& $19.47$ & $96.02$ & $55.29$ & $86.18$ & $18.89$ & $95.86$ & $90.29$ \\
\texttt{DivideMix} 
& $79.28$ & $78.13$ 
& $68.73$ & $80.40$ 
& $55.05$ & $86.76$ 
& $26.58$ & $94.70$ 
& $51.37$ & $87.91$ 
& $74.59$ & $77.40$ 
& $59.27$ & $84.22$ 
& $78.30$ \\
\texttt{Co-teaching}
& $51.32$ & $81.34$ & $99.87$ & $32.07$ & $74.58$ & $69.89$ & $94.74$ & $62.07$
& $52.39$ & $81.73$ & $92.63$ & $49.46$ & $77.59$ & $62.76$ & $85.99$ \\

\texttt{CM}
& $1.83$ & $99.62$ & $16.84$ & $97.00$ & $17.64$ & $96.72$ & $18.08$ & $96.81$
& $27.96$ & $94.45$ & $59.38$ & $85.66$ & $23.62$ & $95.04$ & $90.29$ \\
\texttt{VolMinNet}
& $3.95$ & $99.25$ & $15.59$ & $97.29$ & $8.74$ & $98.36$ & $8.31$ & $98.42$
& $22.82$ & $95.84$ & $53.37$ & $87.74$ & $18.80$ & $96.15$ & $94.35$ \\
\midrule

\texttt{NOODLE}
& $1.22$ & $99.77$ & $12.36$ & $97.84$ & $2.86$ & $99.43$ & $11.00$ & $98.04$
& $14.24$ & $97.35$ & $55.58$ & $86.68$ & $16.21$ & $96.52$ & $89.64$ \\
\midrule
\multicolumn{16}{c}{\textbf{Noise = random3}} \\
\midrule
\texttt{KNN}
& $8.79$ & $98.41$ & $29.45$ & $93.81$ & $12.45$ & $97.66$ & $28.16$ & $94.63$
& $25.80$ & $94.26$ & $58.99$ & $85.76$ & $27.27$ & $94.09$ & $87.77$ \\
\texttt{MSP}
& $50.92$ & $90.23$ & $44.39$ & $89.13$ & $25.79$ & $95.34$ & $69.78$ & $81.85$
& $67.02$ & $81.86$ & $65.51$ & $81.76$ & $53.90$ & $86.70$ & $87.77$\\
\texttt{ODIN}
& $41.84$ & $90.63$ & $30.77$ & $92.45$ & $8.82$ & $98.25$ & $42.48$ & $89.24$
& $60.35$ & $77.22$ & $67.20$ & $76.41$ & $41.91$ & $87.37$ & $87.77$ \\
\texttt{Energy}
& $40.77$ & $92.63$ & $26.49$ & $93.06$ & $8.85$ & $98.19$ & $57.99$ & $85.82$
& $64.08$ & $80.66$ & $58.27$ & $83.06$ & $42.74$ & $88.90$ & $87.77$\\
\texttt{ReAct}
& $77.62$ & $77.76$ & $34.33$ & $92.52$ & $16.16$ & $96.96$ & $67.01$ & $79.29$
& $85.32$ & $62.89$ & $66.42$ & $77.99$ & $57.81$ & $81.24$ & $87.77$ \\
\texttt{Mahalanobis}
& $7.87$ & $98.31$ & $26.49$ & $94.43$ & $7.01$ & $98.64$ & $54.91$ & $82.04$
& $28.95$ & $90.66$ & $87.15$ & $57.60$ & $35.40$ & $86.95$ & $87.77$\\
\texttt{CIDER}
& $98.96$ & $18.22$ & $98.75$ & $24.24$ & $75.41$ & $69.92$ & $91.38$ & $57.40$
& $88.28$ & $38.09$ & $93.45$ & $50.68$ & $91.04$ & $43.09$ & $23.21$ \\
\texttt{SSD+}
& $99.54$ & $26.49$ & $99.02$ & $42.89$ & $100.00$ & $33.08$ & $100.00$ & $33.42$
& $98.54$ & $17.58$ & $99.15$ & $43.08$ & $99.38$ & $32.76$ & $19.20$ \\
\texttt{SNN}
& $12.97$ & $97.67$ & $34.27$ & $91.84$ & $16.04$ & $96.81$ & $53.01$ & $89.46$
& $29.40$ & $92.96$ & $59.90$ & $83.95$ & $34.26$ & $92.12$ & $87.94$\\
\texttt{SCE}
& $6.08$ & $98.81$ & $16.65$ & $96.94$ & $13.47$ & $96.64$ & $24.29$ & $95.55$
& $23.32$ & $95.23$ & $65.06$ & $83.22$ & $24.81$ & $94.40$ & $89.39$ \\
\texttt{GCE}
& $7.30$ & $98.63$ & $17.23$ & $96.88$ & $5.28$ & $98.96$ & $10.88$ & $97.86$
& $20.04$ & $95.84$ & $57.98$ & $85.49$ & $19.78$ & $95.61$ & $90.71$ \\

\texttt{DivideMix} 
& $26.15$ & $95.47$ 
& $57.09$ & $89.78$ 
& $38.77$ & $92.91$ 
& $46.40$ & $89.32$ 
& $49.93$ & $87.99$ 
& $74.53$ & $80.78$ 
& $48.81$ & $89.38$ 
& $69.20$ \\

\texttt{Co-teaching}
& $50.65$ & $80.80$ & $99.96$ & $23.83$ & $99.55$ & $56.65$ & $95.49$ & $58.16$
& $52.45$ & $80.56$ & $93.54$ & $48.86$ & $81.94$ & $58.15$ & $86.53$ \\

\texttt{CM}
& $2.56$ & $99.48$ & $12.00$ & $97.75$ & $6.82$ & $98.60$ & $23.46$ & $95.99$
& $23.74$ & $94.94$ & $55.65$ & $86.32$ & $20.70$ & $95.51$ & $90.79$ \\
\texttt{VolMinNet}
& $9.80$ & $98.06$ & $10.10$ & $98.02$ & $7.61$ & $98.40$ & $24.25$ & $95.63$
& $25.74$ & $94.40$ & $55.65$ & $85.58$ & $22.19$ & $95.02$ &  $94.36$\\
\midrule
\texttt{NOODLE}
& $2.16$ & $99.55$ & $13.31$ & $97.52$ & $4.11$ & $99.15$ & $5.26$ & $98.79$
& $18.26$ & $95.86$ & $53.31$ & $87.46$ & $16.07$ & $96.39$ & $90.32$ \\
\bottomrule
\end{tabular}
}
\end{table*}

\begin{table*}[h]
\centering
\caption{OOD detection performance (FPR95$\downarrow$ / AUROC$\uparrow$) on Animal-10N with real noisy labels using a DenseNet-100 encoder.}
\label{tab:animal10n-detail}
\setlength{\tabcolsep}{3.5pt}
\renewcommand{\arraystretch}{1.2}
\resizebox{\textwidth}{!}{
\begin{tabular}{lccccccccccccccc}
\toprule
\textbf{Method}
& \multicolumn{2}{c}{SVHN}
& \multicolumn{2}{c}{FashionMNIST}
& \multicolumn{2}{c}{LSUN}
& \multicolumn{2}{c}{iSUN}
& \multicolumn{2}{c}{DTD}
& \multicolumn{2}{c}{Places365}
& \multicolumn{2}{c}{Average} 
& \textbf{ID Acc.} \\
\cmidrule(lr){2-3}\cmidrule(lr){4-5}\cmidrule(lr){6-7}\cmidrule(lr){8-9}\cmidrule(lr){10-11}\cmidrule(lr){12-13}\cmidrule(lr){14-15}\cmidrule(lr){16-16}
& FPR95$\downarrow$ & AUROC$\uparrow$
& FPR95$\downarrow$ & AUROC$\uparrow$
& FPR95$\downarrow$ & AUROC$\uparrow$
& FPR95$\downarrow$ & AUROC$\uparrow$
& FPR95$\downarrow$ & AUROC$\uparrow$
& FPR95$\downarrow$ & AUROC$\uparrow$
& FPR95$\downarrow$ & AUROC$\uparrow$
& \\
\midrule

\texttt{KNN}         & $45.29$ & $89.55$ & $58.05$ & $90.30$ & $66.19$ & $79.93$ & $90.66$ & $62.18$ & $73.17$ & $74.54$ & $89.26$ & $65.71$ & $70.44$ & $77.04$ & $81.52$ \\
\texttt{MSP}         & $93.12$ & $64.84$ & $82.74$ & $79.61$ & $84.00$ & $74.53$ & $96.60$ & $34.67$ & $96.97$ & $41.18$ & $90.40$ & $64.56$ & $90.64$ & $59.90$ & $81.52$ \\
\texttt{ODIN}        & $79.29$ & $64.67$ & $51.20$ & $88.26$ & $48.52$ & $87.84$ & $96.52$ & $34.97$ & $95.43$ & $36.66$ & $90.85$ & $62.89$ & $76.97$ & $62.55$ & $81.52$ \\
\texttt{Energy}      & $81.91$ & $74.54$ & $66.32$ & $88.64$ & $31.94$ & $94.83$ & $96.32$ & $45.16$ & $89.02$ & $69.24$ & $88.09$ & $74.71$ & $75.60$ & $74.52$ & $81.52$ \\
\texttt{ReAct}       & $79.18$ & $77.62$ & $72.85$ & $85.98$ & $42.96$ & $92.02$ & $95.69$ & $58.85$ & $91.88$ & $52.46$ & $91.43$ & $59.99$ & $79.00$ & $71.15$ & $81.52$ \\

\texttt{Mahalanobis} & $31.73$ & $91.49$ & $97.47$ & $42.44$ & $68.73$ & $66.87$ & $4.87$  & $98.87$ & $33.39$ & $86.64$ & $91.04$ & $51.67$ & $54.54$ & $73.00$ & $81.52$ \\
\texttt{CIDER} 
& $98.86$ & $39.21$ 
& $98.50$ & $78.80$ 
& $96.32$ & $61.83$ 
& $97.93$ & $37.85$ 
& $99.89$ & $20.98$ 
& $99.19$ & $39.05$ 
& $98.44$ & $39.78$ & $20.24$ \\

\texttt{SSD+}
& $74.30$ & $66.08$
& $89.00$ & $57.00$
& $97.69$ & $31.96$
& $66.50$ & $65.96$
& $91.21$ & $43.10$
& $98.31$ & $37.19$
& $85.60$ & $48.86$ & $19.60$ \\

\texttt{SNN}         & $41.40$ & $91.28$ & $11.58$ & $97.79$ & $12.04$ & $97.63$ & $39.68$ & $91.61$ & $29.22$ & $93.29$ & $54.67$ & $90.27$ & $31.43$ & $93.65$ & $81.52$ \\

\texttt{SCE} 
& $29.06$ & $90.96$ 
& $41.71$ & $93.05$ 
& $27.00$ & $94.63$ 
& $9.45$  & $98.39$ 
& $29.59$ & $94.10$ 
& $55.03$ & $89.71$ 
& $31.97$ & $93.47$ & $ 81.22$ \\
\texttt{GCE} 
& $28.98$ & $91.90$ 
& $43.05$ & $91.28$ 
& $23.75$ & $94.14$ 
& $27.02$ & $95.91$ 
& $30.18$ & $92.52$ 
& $66.74$ & $84.15$ 
& $36.62$ & $91.65$ & $80.86$ \\

\texttt{DivideMix} 
& $16.15$ & $96.44$ 
& $19.84$ & $96.28$ 
& $18.42$ & $95.98$ 
& $60.56$ & $86.34$ 
& $40.14$ & $85.92$ 
& $50.52$ & $89.62$ 
& $34.27$ & $91.77$ & $79.63$ \\

\texttt{Co-teaching} 
& $68.04$ & $83.45$ 
& $99.90$ & $8.68$ 
& $84.54$ & $55.97$ 
& $11.71$ & $97.11$ 
& $49.72$ & $82.39$ 
& $96.92$ & $42.70$ 
& $68.47$ & $61.72$ & $74.08$ \\

\texttt{CM}          & $25.84$ & $95.25$ & $15.41$ & $97.22$ & $13.42$ & $97.44$ & $63.61$ & $84.11$ & $37.06$ & $90.88$ & $45.65$ & $91.58$ & $33.50$ & $92.75$ & $82.48$ \\
\texttt{VolMinNet}   & $15.08$ & $96.79$ & $29.22$ & $94.51$ & $11.80$ & $97.73$ & $40.62$ & $90.67$ & $25.20$ & $94.64$ & $53.63$ & $90.21$ & $29.26$ & $94.09$ & $81.78$ \\
\midrule
\texttt{NOODLE}      & $26.49$ & $94.77$ & $24.41$ & $95.75$ & $11.36$ & $97.86$ & $17.29$ & $96.70$ & $18.21$ & $96.33$ & $53.76$ & $89.37$ & $25.25$ & $95.13$ & $82.98$ \\
\bottomrule
\end{tabular}
}
\end{table*}

\clearpage

    \begin{table*}[h]
\centering
\caption{OOD detection performance (FPR95$\downarrow$ / AUROC$\uparrow$) on CIFAR-100 with real noisy labels using a DenseNet-100 encoder.}
\label{tab:cifar100-detail}
\setlength{\tabcolsep}{3.5pt}
\renewcommand{\arraystretch}{1.2} 
\resizebox{\textwidth}{!}{
\begin{tabular}{lccccccccccccccc}
\toprule
\textbf{Method}
& \multicolumn{2}{c}{SVHN}
& \multicolumn{2}{c}{FashionMNIST}
& \multicolumn{2}{c}{LSUN}
& \multicolumn{2}{c}{iSUN}
& \multicolumn{2}{c}{DTD}
& \multicolumn{2}{c}{Places365}
& \multicolumn{2}{c}{Average}
& \textbf{ID Acc.} \\
\cmidrule(lr){2-3}\cmidrule(lr){4-5}\cmidrule(lr){6-7}\cmidrule(lr){8-9}\cmidrule(lr){10-11}\cmidrule(lr){12-13}\cmidrule(lr){14-15}\cmidrule(lr){16-16}
& FPR95$\downarrow$ & AUROC$\uparrow$
& FPR95$\downarrow$ & AUROC$\uparrow$
& FPR95$\downarrow$ & AUROC$\uparrow$
& FPR95$\downarrow$ & AUROC$\uparrow$
& FPR95$\downarrow$ & AUROC$\uparrow$
& FPR95$\downarrow$ & AUROC$\uparrow$
& FPR95$\downarrow$ & AUROC$\uparrow$
& \\
\midrule
\texttt{KNN}            & $11.08$ & $97.63$ & $42.68$ & $92.88$ & $28.07$ & $93.32$ & $53.09$ & $81.81$ & $32.73$ & $91.70$ & $91.56$ & $61.90$ & $43.20$ & $86.54$ & $52.48$ \\
\texttt{MSP}            & $86.65$ & $75.40$ & $76.23$ & $81.21$ & $52.72$ & $88.08$ & $88.69$ & $67.64$ & $92.36$ & $60.23$ & $89.81$ & $63.40$ & $81.08$ & $72.66$ & $52.48$ \\
\texttt{ODIN}           & $93.41$ & $68.56$ & $58.06$ & $87.70$ & $26.85$ & $95.10$ & $70.69$ & $83.42$ & $92.78$ & $59.70$ & $88.51$ & $65.36$ & $71.72$ & $76.64$ & $52.48$ \\
\texttt{Energy}         & $97.07$ & $47.12$ & $53.27$ & $91.52$ & $35.35$ & $93.49$ & $90.27$ & $68.41$ & $99.29$ & $3.46$  & $98.31$ & $6.41$  & $78.93$ & $51.74$ &$52.48$\\
\texttt{ReAct}          & $97.07$ & $47.12$ & $53.27$ & $91.52$ & $35.35$ & $93.49$ & $90.27$ & $68.41$ & $97.53$ & $38.90$ & $83.94$ & $66.32$ & $76.24$ & $67.63$ & $52.48$\\
\texttt{Mahalanobis}    & $64.60$ & $82.25$ & $99.38$ & $44.82$ & $95.23$ & $49.37$ & $53.12$ & $83.45$ & $42.13$ & $84.50$ & $96.43$ & $48.47$ & $75.15$ & $65.47$ & $52.48$ \\

\texttt{CIDER}          & $98.86$ & $39.76$ & $99.33$ & $29.09$ & $99.19$ & $39.05$ & $96.36$ & $61.84$ & $97.93$ & $37.85$ & $99.89$ & $20.99$ & $98.59$ & $38.10$ & $19.76$ \\
\texttt{SSD+}           & $99.00$ & $39.35$ & $99.33$ & $29.07$ & $99.19$ & $39.05$ & $96.36$ & $61.84$ & $97.93$ & $37.85$ & $99.89$ & $20.99$ & $98.62$ & $38.03$ & $ 15.56$ \\

\texttt{SNN}            & $16.78$ & $96.78$ & $35.98$ & $93.88$ & $91.30$ & $61.59$ & $56.80$ & $84.25$ & $30.82$ & $92.35$ & $27.24$ & $93.93$ & $43.15$ & $87.13$ & $58.06$ \\
\texttt{SCE}            & $16.87$ & $96.64$ & $14.13$ & $97.12$ & $30.67$ & $91.97$ & $75.99$ & $62.21$ & $50.27$ & $86.01$ & $88.84$ & $64.92$ & $46.13$ & $83.15$ & $ 60.74$ \\
\texttt{GCE}            & $63.18$ & $82.35$ & $58.27$ & $88.42$ & $63.09$ & $77.88$ & $80.78$ & $66.52$ & $58.03$ & $82.63$ & $87.90$ & $67.42$ & $68.54$ & $77.54$ & $58.21$ \\
\texttt{DivideMix}      & $30.53$ & $94.00$ & $37.22$ & $93.63$ & $67.81$ & $83.31$ & $67.09$ & $70.49$ & $48.55$ & $85.01$ & $86.47$ & $71.10$ & $56.28$ & $82.92$ & $33.26$ \\
\texttt{Co-teaching}    & $51.77$ & $83.17$ & $99.96$ & $27.69$ & $85.42$ & $66.48$ & $98.34$ & $47.85$ & $59.41$ & $79.01$ & $95.20$ & $53.47$ & $81.68$ & $59.61$ & $46.18$\\
\texttt{CM}             & $35.24$ & $92.09$ & $40.79$ & $92.55$ & $36.79$ & $90.95$ & $54.59$ & $79.49$ & $40.46$ & $89.36$ & $89.26$ & $68.02$ & $49.52$ & $85.41$ & $59.06$ \\
\texttt{VolMinNet}      & $36.03$ & $91.87$ & $50.23$ & $91.22$ & $45.13$ & $88.00$ & $74.76$ & $61.76$ & $43.26$ & $88.84$ & $90.49$ & $66.66$ & $56.65$ & $81.39$ &$60.10$ \\
\midrule
\texttt{NOODLE}         & $44.68$ & $91.36$ & $22.33$ & $96.32$ & $1.78$  & $99.47$ & $23.73$ & $94.97$ & $38.12$ & $89.17$ & $88.61$ & $65.18$ & $\textbf{36.54}$ & $89.41$ &$60.89$ \\
\bottomrule
\end{tabular}
}
\end{table*}

\clearpage

\newpage
\end{document}

%% file: preamble.tex
\newcommand{\X}{\boldsymbol{X}}

\newcommand{\x}{\boldsymbol{x}}

\newcommand{\T}{{\!\top\!}}

\newcommand{\tX}{\underline{\bm X}}

\usepackage{calc,amssymb,bm,url,color,theorem,epstopdf,nicefrac, booktabs}


%% file: neurips_2025.bbl
\begin{thebibliography}{10}

\bibitem{goodfellow2014explaining}
Ian~J Goodfellow, Jonathon Shlens, and Christian Szegedy.
\newblock Explaining and harnessing adversarial examples.
\newblock {\em arXiv preprint arXiv:1412.6572}, 2014.

\bibitem{geiger2012we}
Andreas Geiger, Philip Lenz, and Raquel Urtasun.
\newblock Are we ready for autonomous driving? the kitti vision benchmark suite.
\newblock In {\em 2012 IEEE conference on computer vision and pattern recognition}, pages 3354--3361. IEEE, 2012.

\bibitem{thomas2017unsupervised}
Thomas Schlegl, Philipp Seeb{\"o}ck, Sebastian~M. Waldstein, Ursula Schmidt-Erfurth, and Georg Langs.
\newblock Unsupervised anomaly detection with generative adversarial networks to guide marker discovery.
\newblock In {\em Information Processing in Medical Imaging}, pages 146--157. Springer International Publishing, 2017.

\bibitem{hendrycks2016baseline}
Dan Hendrycks and Kevin Gimpel.
\newblock A baseline for detecting misclassified and out-of-distribution examples in neural networks.
\newblock {\em arXiv preprint arXiv:1610.02136}, 2016.

\bibitem{yang2024generalized}
Jingkang Yang, Kaiyang Zhou, Yixuan Li, and Ziwei Liu.
\newblock Generalized out-of-distribution detection: A survey.
\newblock {\em International Journal of Computer Vision}, 132(12):5635--5662, 2024.

\bibitem{liang2018enhancing}
Shiyu Liang, Yixuan Li, and R.~Srikant.
\newblock Enhancing the reliability of out-of-distribution image detection in neural networks.
\newblock In {\em International Conference on Learning Representations (ICLR)}, 2018.

\bibitem{hendrycks2022scaling}
Dan Hendrycks, Steven Basart, Mantas Mazeika, Andy Zou, Joseph Kwon, Mohammadreza Mostajabi, Jacob Steinhardt, and Dawn Song.
\newblock Scaling out-of-distribution detection for real-world settings.
\newblock In {\em Proceedings of the 39th International Conference on Machine Learning (ICML)}, pages 8759--8773. PMLR, 2022.

\bibitem{sun2022dice}
Yiyou Sun and Yixuan Li.
\newblock Dice: Leveraging sparsification for out-of-distribution detection.
\newblock In {\em Computer Vision – ECCV 2022}, pages 691--708. Springer Nature Switzerland, 2022.

\bibitem{sun2021react}
Yiyou Sun, Chuan Guo, and Yixuan Li.
\newblock React: Out-of-distribution detection with rectified activations.
\newblock In {\em Advances in Neural Information Processing Systems (NeurIPS)}, pages 144--157. Curran Associates, Inc., 2021.

\bibitem{dong2022neural}
Xin Dong, Junfeng Guo, Ang Li, Wei-Te Ting, Cong Liu, and HT~Kung.
\newblock Neural mean discrepancy for efficient out-of-distribution detection.
\newblock In {\em Proceedings of the IEEE/CVF Conference on Computer Vision and Pattern Recognition}, pages 19217--19227, 2022.

\bibitem{lee2018simple}
Kimin Lee, Kibok Lee, Honglak Lee, and Jinwoo Shin.
\newblock A simple unified framework for detecting out-of-distribution samples and adversarial attacks.
\newblock In {\em Advances in Neural Information Processing Systems (NeurIPS)}. Curran Associates, Inc., 2018.

\bibitem{sun2022deepnn}
Yiyou Sun, Yifei Ming, Xiaojin Zhu, and Yixuan Li.
\newblock Out-of-distribution detection with deep nearest neighbors.
\newblock In {\em Proceedings of the 39th International Conference on Machine Learning (ICML)}, pages 20827--20840. PMLR, 2022.

\bibitem{ming2023hypersphericalood}
Yifei Ming, Yiyou Sun, Ousmane Dia, and Yixuan Li.
\newblock How to exploit hyperspherical embeddings for out-of-distribution detection?, 2023.

\bibitem{SehwagSSD}
Vikash Sehwag, Mung Chiang, and Prateek Mittal.
\newblock {SSD:} {A} unified framework for self-supervised outlier detection.
\newblock {\em CoRR}, abs/2103.12051, 2021.

\bibitem{ghosal2023overcomeood}
Soumya~Suvra Ghosal, Yiyou Sun, and Yixuan Li.
\newblock How to overcome curse-of-dimensionality for out-of-distribution detection?
\newblock {\em Proceedings of the AAAI Conference on Artificial Intelligence}, 38(18):19849--19857, Mar. 2024.

\bibitem{buhrmester2016amazon}
Michael Buhrmester, Tracy Kwang, and Samuel~D Gosling.
\newblock Amazon's mechanical turk: A new source of inexpensive, yet high-quality data?
\newblock 2016.

\bibitem{humblot2024noisy}
Galadrielle Humblot-Renaux, Sergio Escalera, and Thomas~B Moeslund.
\newblock A noisy elephant in the room: Is your out-of-distribution detector robust to label noise?
\newblock In {\em Proceedings of the IEEE/CVF Conference on Computer Vision and Pattern Recognition}, pages 22626--22636, 2024.

\bibitem{song2022learning}
Hwanjun Song, Minseok Kim, Dongmin Park, Yooju Shin, and Jae-Gil Lee.
\newblock Learning from noisy labels with deep neural networks: A survey.
\newblock {\em IEEE transactions on neural networks and learning systems}, 34(11):8135--8153, 2022.

\bibitem{arpit2017closer}
Devansh Arpit, Stanis\l{}aw Jastrzundefinedbski, Nicolas Ballas, David Krueger, Emmanuel Bengio, Maxinder~S. Kanwal, Tegan Maharaj, Asja Fischer, Aaron Courville, Yoshua Bengio, and Simon Lacoste-Julien.
\newblock A closer look at memorization in deep networks.
\newblock In {\em Proceedings of International Conference on Machine Learning}, page 233–242, 2017.

\bibitem{zhang2017understanding}
Chiyuan Zhang, Samy Bengio, Moritz Hardt, Benjamin Recht, and Oriol Vinyals.
\newblock Understanding deep learning requires rethinking generalization.
\newblock In {\em Proceedings of International Conference on Learning Representations}, 2016.

\bibitem{liu2016classification}
Tongliang Liu and Dacheng Tao.
\newblock Classification with noisy labels by importance reweighting.
\newblock {\em IEEE Transactions on Pattern Analysis and Machine Intelligence}, 38:447--461, 2016.

\bibitem{patrini2017making}
Giorgio Patrini, Alessandro Rozza, Aditya Krishna~Menon, Richard Nock, and Lizhen Qu.
\newblock Making deep neural networks robust to label noise: A loss correction approach.
\newblock In {\em Proceedings of the IEEE Conference on Computer Vision and Pattern Recognition (CVPR)}, July 2017.

\bibitem{li2021provably}
Xuefeng Li, Tongliang Liu, Bo~Han, Gang Niu, and Masashi Sugiyama.
\newblock Provably end-to-end label-noise learning without anchor points.
\newblock In {\em Proceedings of International Conference on Machine Learning}, pages 6403--6413, 2021.

\bibitem{xia2020part}
Xiaobo Xia, Tongliang Liu, Bo~Han, Nannan Wang, Mingming Gong, Haifeng Liu, Gang Niu, Dacheng Tao, and Masashi Sugiyama.
\newblock Part-dependent label noise: Towards instance-dependent label noise.
\newblock In {\em Advances in Neural Information Processing Systems}, volume~33, pages 7597--7610, 2020.

\bibitem{Yang2021EstimatingIB}
Shuo Yang, Erkun Yang, Bo~Han, Yang Liu, Min Xu, Gang Niu, and Tongliang Liu.
\newblock Estimating instance-dependent {B}ayes-label transition matrix using a deep neural network.
\newblock In {\em Proceedings of International Conference on Machine Learning}, 2021.

\bibitem{cheng2020learning}
Jiacheng Cheng, Tongliang Liu, Kotagiri Ramamohanarao, and Dacheng Tao.
\newblock Learning with bounded instance and label-dependent label noise.
\newblock In {\em Proceedings of International Conference on Machine Learning}, volume 119, pages 1789--1799, 2020.

\bibitem{zhang2018generalized}
Zhilu Zhang and Mert Sabuncu.
\newblock Generalized cross entropy loss for training deep neural networks with noisy labels.
\newblock {\em Advances in neural information processing systems}, 31, 2018.

\bibitem{lyu2019curriculum}
Yueming Lyu and Ivor~W Tsang.
\newblock Curriculum loss: Robust learning and generalization against label corruption.
\newblock {\em arXiv preprint arXiv:1905.10045}, 2019.

\bibitem{wang2019symmetric}
Yisen Wang, Xingjun Ma, Zaiyi Chen, Yuan Luo, Jinfeng Yi, and James Bailey.
\newblock Symmetric cross entropy for robust learning with noisy labels.
\newblock In {\em Proceedings of the IEEE/CVF international conference on computer vision}, pages 322--330, 2019.

\bibitem{jiang2018mentornet}
Lu~Jiang, Zhengyuan Zhou, Thomas Leung, Li-Jia Li, and Li~Fei-Fei.
\newblock Mentornet: Learning data-driven curriculum for very deep neural networks on corrupted labels.
\newblock In {\em International conference on machine learning}, pages 2304--2313. PMLR, 2018.

\bibitem{yu2019does}
Xingrui Yu, Bo~Han, Jiangchao Yao, Gang Niu, Ivor Tsang, and Masashi Sugiyama.
\newblock How does disagreement help generalization against label corruption?
\newblock In {\em International conference on machine learning}, pages 7164--7173. PMLR, 2019.

\bibitem{nguyen2019self}
Duc~Tam Nguyen, Chaithanya~Kumar Mummadi, Thi Phuong~Nhung Ngo, Thi Hoai~Phuong Nguyen, Laura Beggel, and Thomas Brox.
\newblock Self: Learning to filter noisy labels with self-ensembling.
\newblock {\em arXiv preprint arXiv:1910.01842}, 2019.

\bibitem{han2018co}
Bo~Han, Quanming Yao, Xingrui Yu, Gang Niu, Miao Xu, Weihua Hu, Ivor Tsang, and Masashi Sugiyama.
\newblock Co-teaching: Robust training of deep neural networks with extremely noisy labels.
\newblock {\em Advances in neural information processing systems}, 31, 2018.

\bibitem{li2020dividemix}
Junnan Li, Richard Socher, and Steven~CH Hoi.
\newblock Dividemix: Learning with noisy labels as semi-supervised learning.
\newblock {\em arXiv preprint arXiv:2002.07394}, 2020.

\bibitem{tanno2019learning}
Ryutaro Tanno, Ardavan Saeedi, Swami Sankaranarayanan, Daniel~C Alexander, and Nathan Silberman.
\newblock Learning from noisy labels by regularized estimation of annotator confusion.
\newblock In {\em Proceedings of the IEEE/CVF conference on computer vision and pattern recognition}, pages 11244--11253, 2019.

\bibitem{ibrahim2023deep}
Shahana Ibrahim, Tri Nguyen, and Xiao Fu.
\newblock Deep learning from crowdsourced labels: Coupled cross-entropy minimization, identifiability, and regularization.
\newblock In {\em Proceedings of International Conference on Learning Representations}, 2023.

\bibitem{candes2011robust}
Emmanuel~J Cand{\`e}s, Xiaodong Li, Yi~Ma, and John Wright.
\newblock Robust principal component analysis?
\newblock {\em Journal of the ACM (JACM)}, 58(3):1--37, 2011.

\bibitem{zhang2011image}
Chunjie Zhang, Jing Liu, Qi~Tian, Changsheng Xu, Hanqing Lu, and Songde Ma.
\newblock Image classification by non-negative sparse coding, low-rank and sparse decomposition.
\newblock In {\em CVPR 2011}, pages 1673--1680. IEEE, 2011.

\bibitem{candes2009rpca}
Emmanuel~J. Candes, Xiaodong Li, Yi~Ma, and John Wright.
\newblock Robust principal component analysis?, 2009.

\bibitem{alternateminialgo}
Yilun Wang, Junfeng Yang, Wotao Yin, and Yin Zhang.
\newblock A new alternating minimization algorithm for total variation image reconstruction.
\newblock {\em SIAM J. Imaging Sciences}, 1:248--272, 01 2008.

\bibitem{rokhlin2010randomized}
Vladimir Rokhlin, Arthur Szlam, and Mark Tygert.
\newblock A randomized algorithm for principal component analysis.
\newblock {\em SIAM Journal on Matrix Analysis and Applications}, 31(3):1100--1124, 2010.

\bibitem{gu2015subspace}
Ming Gu.
\newblock Subspace iteration randomization and singular value problems.
\newblock {\em SIAM Journal on Scientific Computing}, 37(3):A1139--A1173, 2015.

\bibitem{Krizhevsky2009LearningML}
Alex Krizhevsky.
\newblock Learning multiple layers of features from tiny images.
\newblock 2009.

\bibitem{wei2022learning}
Jiaheng Wei, Zhaowei Zhu, Hao Cheng, Tongliang Liu, Gang Niu, and Yang Liu.
\newblock Learning with noisy labels revisited: A study using real-world human annotations.
\newblock In {\em International Conference on Learning Representations}, 2022.

\bibitem{song2019selfie}
Hwanjun Song, Minseok Kim, and Jae-Gil Lee.
\newblock {SELFIE}: Refurbishing unclean samples for robust deep learning.
\newblock In {\em ICML}, 2019.

\bibitem{netzer}
Yuval Netzer, Tao Wang, Adam Coates, Alessandro Bissacco, Bo~Wu, and Andrew~Y. Ng.
\newblock Reading digits in natural images with unsupervised feature learning.
\newblock In {\em NIPS Workshop on Deep Learning and Unsupervised Feature Learning 2011}, 2011.

\bibitem{XiaoFahionmnist}
Han Xiao, Kashif Rasul, and Roland Vollgraf.
\newblock Fashion-mnist: a novel image dataset for benchmarking machine learning algorithms.
\newblock {\em CoRR}, abs/1708.07747, 2017.

\bibitem{yu2016lsun}
Fisher Yu, Ari Seff, Yinda Zhang, Shuran Song, Thomas Funkhouser, and Jianxiong Xiao.
\newblock Lsun: Construction of a large-scale image dataset using deep learning with humans in the loop, 2016.

\bibitem{panisun}
Junting Pan and Xavier Gir{\'{o}}{-}i{-}Nieto.
\newblock End-to-end convolutional network for saliency prediction.
\newblock {\em CoRR}, abs/1507.01422, 2015.

\bibitem{Cimpoi}
Mircea Cimpoi, Subhransu Maji, Iasonas Kokkinos, Sammy Mohamed, and Andrea Vedaldi.
\newblock Describing textures in the wild.
\newblock {\em CoRR}, abs/1311.3618, 2013.

\bibitem{zhou2016}
Bolei Zhou, Aditya Khosla, {\`{A}}gata Lapedriza, Antonio Torralba, and Aude Oliva.
\newblock Places: An image database for deep scene understanding.
\newblock {\em CoRR}, abs/1610.02055, 2016.

\bibitem{hendrycks2017baseline}
Dan Hendrycks and Kevin Gimpel.
\newblock A baseline for detecting misclassified and out-of-distribution examples in neural networks.
\newblock In {\em 5th International Conference on Learning Representations, {ICLR} 2017, Toulon, France, April 24-26, 2017, Conference Track Proceedings}. OpenReview.net, 2017.

\bibitem{Liu2020EnergybasedOD}
Weitang Liu, Xiaoyun Wang, John~Douglas Owens, and Yixuan Li.
\newblock Energy-based out-of-distribution detection.
\newblock {\em ArXiv}, abs/2010.03759, 2020.

\bibitem{khosla2020supervised}
Prannay Khosla, Piotr Teterwak, Chen Wang, Aaron Sarna, Yonglong Tian, Phillip Isola, Aaron Maschinot, Ce~Liu, and Dilip Krishnan.
\newblock Supervised contrastive learning.
\newblock {\em Advances in neural information processing systems}, 33:18661--18673, 2020.

\bibitem{Li2021ProvablyEL}
Xuefeng Li, Tongliang Liu, Bo~Han, Gang Niu, and Masashi Sugiyama.
\newblock Provably end-to-end label-noise learning without anchor points.
\newblock In {\em International Conference on Machine Learning}, 2021.

\bibitem{wang2019symmetriccrossentropyrobust}
Yisen Wang, Xingjun Ma, Zaiyi Chen, Yuan Luo, Jinfeng Yi, and James Bailey.
\newblock Symmetric cross entropy for robust learning with noisy labels, 2019.

\bibitem{zhang2018generalizedcrossentropyloss}
Zhilu Zhang and Mert~R. Sabuncu.
\newblock Generalized cross entropy loss for training deep neural networks with noisy labels, 2018.

\bibitem{li2020dividemixlearningnoisylabels}
Junnan Li, Richard Socher, and Steven C.~H. Hoi.
\newblock Dividemix: Learning with noisy labels as semi-supervised learning, 2020.

\bibitem{han2018coteachingrobusttrainingdeep}
Bo~Han, Quanming Yao, Xingrui Yu, Gang Niu, Miao Xu, Weihua Hu, Ivor Tsang, and Masashi Sugiyama.
\newblock Co-teaching: Robust training of deep neural networks with extremely noisy labels, 2018.

\bibitem{densenet}
Gao Huang, Zhuang Liu, and Kilian~Q. Weinberger.
\newblock Densely connected convolutional networks.
\newblock {\em CoRR}, abs/1608.06993, 2016.

\end{thebibliography}
